\documentclass[11pt, a4paper, logo, twocolumn, internal, copyright, nonumbering]{googledeepmind}

\usepackage[authoryear, sort&compress, round]{natbib}
\usepackage{microtype}
\usepackage{graphicx}
\usepackage{booktabs} 

\usepackage{hyperref}

\usepackage{xcolor}

\usepackage{amsmath}
\usepackage{amssymb}
\usepackage{mathtools}
\usepackage{amsthm}
\usepackage{booktabs}
\usepackage{multirow}
\usepackage{url}            
\usepackage{amsfonts}       
\usepackage{nicefrac}       
\usepackage{xcolor}         
\usepackage{algorithm}
\usepackage{physics}

\usepackage{xspace}

\usepackage{array}
\usepackage{arydshln}
\usepackage{wrapfig}
\usepackage{rotating}

\usepackage[capitalize,noabbrev]{cleveref}

\theoremstyle{plain}

\theoremstyle{definition}

\theoremstyle{remark}

\usepackage[textsize=tiny]{todonotes}
\usepackage{subcaption}
\newcommand{\alg}{\ensuremath{{\rm MatQuant}}\xspace}
\newcommand{\spalg}{\ensuremath{{\rm Single\text{ } Precison\text{ }MatQuant}}\xspace}
\newcommand{\epalg}{\ensuremath{{\rm Extra\text{ } Precison\text{ }MatQuant}}\xspace}

\title{Matryoshka Quantization}

\reportnumber{001} 

\renewcommand{\today}{2025-02-05} 

\author[*,1]{Pranav Nair}
\author[*,1]{Puranjay Datta}
\author[1]{Jeff Dean}
\author[1]{Prateek Jain}
\author[1]{Aditya Kusupati}
\affil[1]{Google DeepMind} 
\affil[*]{Equal contribution}

\begin{abstract}

Quantizing model weights is critical for reducing the communication and inference costs of large models. However, quantizing models -- especially to low precisions like int4 or int2 -- requires a trade-off in model quality; int2, in particular, is known to severely degrade model quality. Consequently, practitioners are often forced to maintain multiple models with different quantization levels or serve a single model that best satisfies the quality-latency trade-off. On the other hand, integer data types, such as int8, inherently possess a nested (Matryoshka) structure where smaller bit-width integers, like int4 or int2, are nested within the most significant bits. Leveraging this insight, in this paper, we propose Matryoshka Quantization (\alg), a novel multi-scale quantization technique that alleviates the aforementioned challenge. This technique allows us to train and maintain a single quantized model but serve it with the precision demanded by the deployment. Furthermore, leveraging \alg's co-training and co-distillation regularization, int2 precision models extracted by \alg outperform standard int2 quantization by up to to 4\% and 7\% with OmniQuant and QAT as base algorithms respectively. Finally, we demonstrate that by using an extra bit to represent outliers, a model with an effective precision of 2.05-bit gives an additional 6\% improvement with OmniQuant as the base algorithm.

\end{abstract}

\begin{document}

\maketitle

\section{Introduction}
\label{sec:intro}
        



\newlength{\subcolumnwidth}
\newenvironment{subcolumns}[1][\columnwidth]
 {\valign\bgroup\hsize=#1\setlength{\subcolumnwidth}{\hsize}\vfil##\vfil\cr}
 {\crcr\egroup}
\newcommand{\nextsubcolumn}[1][]{%
  \cr\noalign{\hfill}
  \if\relax\detokenize{#1}\relax\else\hsize=#1\setlength{\subcolumnwidth}{\hsize}\fi
}
\newcommand{\nextsubfigure}{\vfill}








\begin{figure*}[htp]
\begin{subcolumns}[0.3\textwidth]
\centering
 \vspace{-2mm}\subfloat{{\includegraphics[width=0.4\columnwidth]{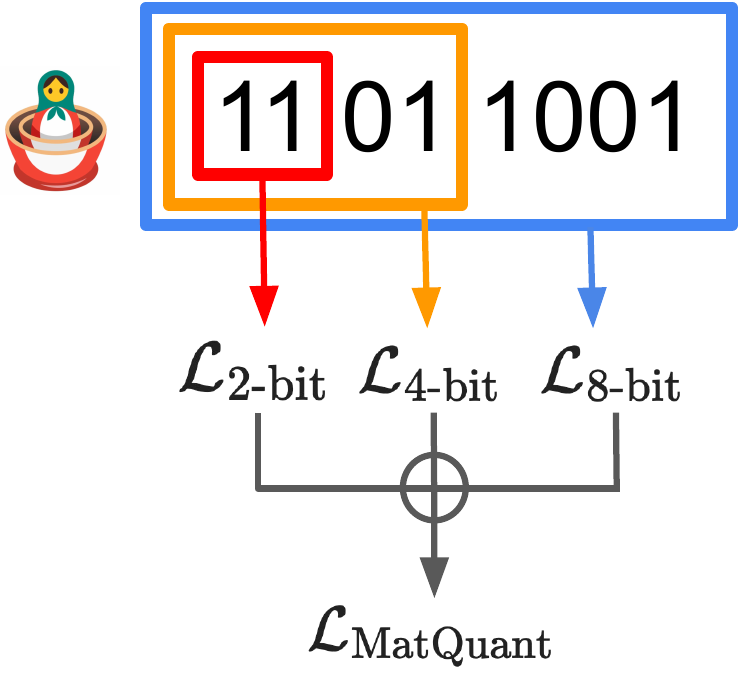}}} \\
 \caption*{\hspace{2.5cm}(a)}
 \hspace*{-1cm}
\nextsubfigure
\vspace{-1mm}
  \subfloat{{\includegraphics[width=0.6\columnwidth]{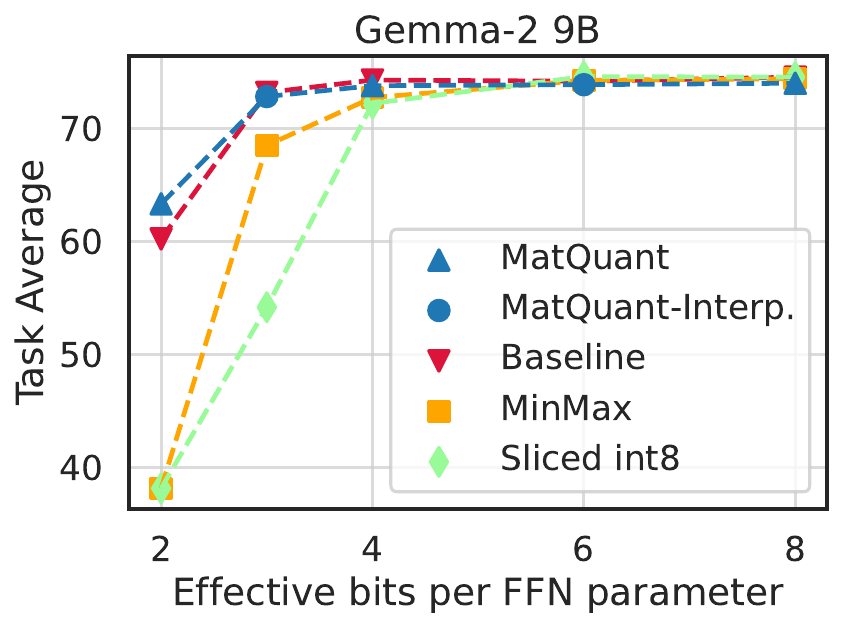}}} \\
  \caption*{\hspace{2.5cm}(b)}
  \hspace*{-1cm}
\nextsubcolumn
\hspace{-64mm}
  \subfloat{{\includegraphics[width=1.40\columnwidth]{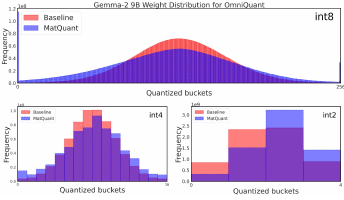}}} \\
  \caption*{\hspace{-0.75cm}(c)}
\end{subcolumns}
\vspace{-1mm}
\caption{(a) \alg is a multi-scale quantization training technique using the inherent Matryoshka structure of int8 $\to$ int4 $\to$ int2. (b) Empirical gains of \alg on downstream tasks, especially $>8\%$ for int2, on Gemma-2 9B with OmniQuant. (c) The right-shifted quantized weight distribution as a consequence of \alg's training mechanism that maximises accuracies across all precisions.}
\label{fig:teaser}
\end{figure*}
Due to their impressive performance, there is a strong push to deploy deep learning models, particularly large language models (LLMs)~\citep{team2024gemini,dubey2024llama,achiam2023gpt} in a large number of scenarios. Due to auto-regressive nature of LLMs, decode latency tends to dominate inference cost. Decode latency itself is dominated by communication cost of transferring model weights from high-bandwidth memory (HBM) to the SRAM or due to transferring weights/activations in a distributed cluster. 


Quantizing weights and/or activations can significantly reduce the overall communication load and is, therefore, one of the most popular techniques for reducing inference costs~\citep{dettmers2022gpt3}. While floating-point representations are standard for training, integer data types such as int8, int4, and int2 are appealing alternatives for inference. However, current methods for quantizing to these varying integer precisions typically treat each target precision as an independent optimization problem, leading to a collection of distinct models rather than a single, versatile one. Furthermore, quantizing to extremely low precisions like int2 is known to be highly inaccurate. In this work, we pose the question of whether both of the above challenges can be addressed; that is, can we train a single model from which we can extract multiple accurate lower-precision models? We answer this question in the affirmative by introducing Matryoshka Quantization (\alg), a novel multi-scale training method that leverages the inherent nested (Matryoshka) structure~\citep{kusupati2022matryoshka} within integer data types (Figure~\ref{fig:teaser}a). Specifically, \textit{slicing} the most significant bits (MSBs) of an int8-quantized weight can directly yield an int4 or int2 model. Existing quantization techniques often neglect this structure, which limits the potential for multi-scale adaptable models operating at various bit-widths with optimal performance.


Instead, \alg simultaneously optimizes model weights across multiple precision levels (e.g., int8, int4, int2). At a high level, we represent each model parameter at different precision levels using shared MSBs, and then jointly optimize the loss for each precision level. This allows us to develop a single quantized model that can effectively operate at any of the chosen bit-widths, offering a spectrum of accuracy-vs-cost options. \alg is a general-purpose technique, applicable to most learning-based quantization methods, such as Quantization Aware Training (QAT)~\citep{jacob2018quantization} and OmniQuant~\citep{shao2023omniquant}.

We demonstrate the efficacy of \alg when applied to quantizing the Feed-Forward Network (FFN) parameters of standard LLMs (Gemma-2 2B, 9B, and Mistral 7B)~\citep{vaswani2017attention} -- typically, FFN is the main latency block hence the focus on improving the most significant component's latency. Our results show that \alg produces int8 and int4 models with comparable accuracy to independently trained baselines, despite the benefit of shared model parameters. Critically, the int2 models generated by \alg significantly outperform their individually trained counterparts, with $4$\% higher accuracy on downstream tasks (Figure~\ref{fig:teaser}b). We also extend \alg to quantize all weights of a Transformer layer.  In Figure~\ref{fig:teaser}c, we find that quantizing with \alg shifts the quantized weight distribution toward higher values, contributing to improved int2 performance. Finally, in Section~\ref{sec:errata}, we also demonstrate that using an extra bit to represent outliers significantly boosts the performance for our sliced int2 models.

Beyond improving chosen precision performance, \alg allows for seamless extraction of interpolative bit-widths, such as int6 and int3. \alg also admits a dense accuracy-vs-cost trade-off by enabling layer-wise Mix'n'Match of different precisions. Therefore, even if the hardware only supports int4 and int2, it's possible to serve models at various effective precisions, tailored to the deployment environment. Overall, \alg and its variants present a significant step toward developing multi-scale models with high flexibility and performance, pushing the boundaries of low-bit quantization for efficient LLM inference.

\section{Related Work}
\label{sec:rw}






Model weight quantization is an extremely powerful and prevalent technique for making resource-intensive neural networks suitable for deployment constraints -- especially modern-day LLMs. Quantization algorithms can be categorized as either learning-free or learning-based. Learning-free methods use limited data to calibrate model parameters without relying on gradient descent. Learning-based methods, however, utilize gradient descent to update either model parameters or auxiliary parameters to aid in quantization.

\vspace{-4mm}
\paragraph{Learning-free Quantization Methods.} Naive quantization methods, such as MinMax, absmax, and zero-point quantization, aim to directly map the range of model weights to the target bit-width -- see~\citep{dettmers2022gpt3} for a detailed background. \citet{dettmers2022gpt3} further improved this by identifying the need to handle outliers with higher precision than the rest of the model weights. The core principle of more recent learning-free quantization methods remains similar while improving various aspects of it and using small amounts of data for calibration. For example, GPTQ~\citep{frantar2022gptq} improves upon min-max quantization by iterating over all the coordinates, quantizing them one at a time, and updating the remaining full-precision coordinates to minimize the layer-wise activation reconstruction error. AWQ~\citep{lin2023awq}, SmoothQuant~\citep{xiao2023smoothquant}, and AffineQuant~\citep{ma2024affinequant} scale the weights and activations to reduce outliers, thus making them easier to quantize. QuIP~\citep{chee2024quip}, FrameQuant~\citep{adepu2024framequant}, and QuaRoT~\citep{quarot} multiply the weights and activations by orthonormal matrices before quantizing to reduce the number of outliers. SqueezeLLM~\citep{squeezellm} uses clustering to obtain the optimal buckets for quantization, and CDQuant~\citep{DBLP:cdquant} improves upon GPTQ by greedily choosing the coordinates to descend along. While learning-free methods are inexpensive and work well at higher bit-widths, they are often suboptimal in the low-precision regime, which benefits greatly from learning-based techniques.


\vspace{-2mm}
\paragraph{Learning-based Quantization Methods.} Quantization Aware Training (QAT)~\citep{jacob2018quantization,abdolrashidi2021pareto} is a logical approach to ensure that models are easy to quantize during inference while retaining high accuracy. However, because QAT involves updating all the model parameters, its adoption for LLMs has been limited. Several recent works improve the performance and efficiency of QAT. LLM-QAT~\citep{DBLP:llmqat} and BitDistiller~\citep{DBLP:BitDistiller} enhance QAT with knowledge distillation from the full-precision model. EfficientQAT~\citep{DBLP:efficientqat} minimizes the block-wise reconstruction error before performing end-to-end training. This significantly reduces the time it takes for QAT to converge. On the other hand, some techniques significantly reduce the overhead by learning only the auxiliary parameters, such as scaling factors and zero-points, that aid in quantization instead of updating the actual weight matrices. For example, OmniQuant~\citep{shao2023omniquant} does not update the model parameters; instead, it learns additional scales and shifting parameters (that aid with quantization) through gradient descent over the block-wise reconstruction error and achieves better accuracy than most QAT techniques. Likewise, SpinQuant~\citep{spinquant} uses gradient descent to learn its rotation matrices. This class of learning-based quantization techniques (OmniQuant, SpinQuant, etc.) is widely adopted due to their appeal of achieving QAT-level accuracy at a fraction of the cost.

\vspace{-3mm}
\paragraph{Multi-scale Training.} Training across multiple data scales (resolutions) was heavily popularized in computer vision for both recognition and generation~\citep{adelson1984pyramid,lin2017feature,denton2015deep}. More recently, the paradigm of multi-scale training has shifted to models~\citep{rippel2014learning,yu2018slimmable,kusupati2022matryoshka,devvrit2023matformer}, where the data remains the same, and models of varying capacity, all nested within one large model, are trained jointly. This joint, nested (Matryoshka-style) learning with varying model sizes results in a smooth accuracy-vs-compute trade-off and is beneficial in many downstream applications and real-world deployments. However, the most obvious structure with a nested nature is the bit structure of the integer data type. Given the success of multi-scale training for inputs, outputs, and model weights, it is imperative to explore it further for integer data types, especially in the context of quantization, which aids in the deployment of resource-intensive LLMs. Following this idea, \citet{any_precision_dnn} have successfully trained a single model that can do well at any precision. However, the experiments were limited to ConvNets and small Neural Networks. In this paper, we extend the idea of nested precision to LLMs and show that it indeed works at scale. We also show that, for the first time, our models are quality neutral for intermediate precisions such as int3 and int6 that we never trained for, and densely span the accuracy-vs-bits trade-off. In Section~\ref{sec:spmatquant}, we show that even to train models for a fixed target precision, having loss over the sliced bits of an 8-bit model does better than training a model explicitly for that precision, indicating that \alg is a fundamentally better way to do low-bit quantization.


\section{Matryoshka Quantization}
\label{sec:method}

We introduce \alg, a general-purpose, multi-scale training technique that works seamlessly with popular learning-based quantization methods such as Quantization Aware Training (QAT)~\citep{jacob2018quantization} and OmniQuant~\citep{shao2023omniquant}. As long as the model or auxiliary parameters are optimized with gradient descent, \alg's multi-scale training technique can be used across chosen bit-widths, leveraging the inherent nested structure of integer data types. In this section, we will elaborate on the preliminaries behind QAT and OmniQuant, alongside our novel proposed approach, \alg.

\vspace*{-2mm}
\subsection{Preliminaries}
\subsubsection{Quantization Aware Training}
Quantization Aware Training (QAT) learns a $c$-bit quantized model by optimizing for the end-to-end cross entropy loss using gradient descent. It uses the quantized weights for the forward pass and a straight through estimator (STE)~\citep{bengio2013estimating} to propagate gradients through the quantization operator during the backward pass.

To mathematically formulate QAT, we define MinMax quantization of a real-valued vector $w$ in $c$ bits as follows:
\begin{equation}
\label{eqn:minmax}
\begin{aligned}
Q_{\text{MM}}(w, c) = \text{clamp}\left(\left\lfloor \frac{w}{\alpha} + z\right\rceil, 0, 2^c-1\right) \\
    \alpha = \frac{\max(w) -\min(w)}{2^c-1}, \quad
    z = -\frac{\min(w)}{\alpha}
\end{aligned}
\end{equation}
where $Q_{\text{MM}}(w, c)$ is the $c$-bit quantized version of $w$, $\alpha$ is the scaling factor and $z$ is the zero point.


Let $W_F$ represent weights of a Transformer LLM and let $\mathcal{D}=\{(x_1, y_1), \ldots, (x_N,y_N)\}$ be a labelled dataset where $x_i$ and $y_i$ represent the input and output respectively. With $L_{\text{CE}}$ as the cross entropy loss, the optimization of QAT is:
\begin{equation}
    \min_{W_F} \frac{1}{N}\sum_{i\in [N]}\mathcal{L}_{\text{CE}}\left( F(x_i;Q_{\text{MM}}\left(W_F, c\right)), y_i\right)
\end{equation}
where $F(\cdot)$ represents the LLM's forward pass.

\subsubsection{OmniQuant}
OmniQuant, unlike QAT, does not update the model parameters. Instead, it learns additional scaling and shifting parameters through gradient descent over layer-wise L2 error reconstruction. These auxiliary parameters aid with quantization. Similar to QAT, OmniQuant also uses a straight through estimator during optimization. However, unlike QAT, OmniQuant operates with limited data, making it much more attractive for resource-scarce settings.

OmniQuant adds two learnable scales, $\gamma$ and $\beta$, to MinMax quantization as follows:
\begin{equation}
\label{eqn:omni_quant}
\begin{aligned}
Q_{\text{Omni}}(w, c) = \text{clamp}\left(\left\lfloor \frac{w}{\alpha} + z\right\rceil, 0, 2^c-1\right) \\
    \alpha = \frac{\gamma\cdot\max(w) -\beta\cdot \min(w)}{2^c-1}, \quad
    z = -\frac{\beta\cdot\min(w)}{\alpha}
\end{aligned}
\end{equation}

OmniQuant also adds another set of learnable shifting and scaling parameters to the FFN's affine projections as follows:
\begin{equation}
    \label{eqn:omni_ffn}
    XW + b \rightarrow \left((X - \delta)\oslash s\right)\cdot Q_{\text{Omni}}(W \odot s) + b + \delta \cdot W
\end{equation}
where $X \in \mathbb{R}^{n \times d}$ is the input to the affine transformation, $W \in \mathbb{R}^{d \times d_{\text{o}}}$ is the linear projection associated with the affine transformation, $b \in \mathbb{R}^{d_{\text{o}}}$ is the bias vector, $\delta \in \mathbb{R}^d$ and $s \in \mathbb{R}^d$ are learnable shift and scale parameters respectively.

With the goal of optimizing the layer-wise L2 error (where a layer consists of an Attention block followed by an FFN block), OmniQuant's overall objective can be portrayed
 as follows:
\begin{equation}
    \label{omni_obj}
    \min_{\gamma,\beta,\delta,s} ||F_{l}(W_F^l), X_{l}) - F_{l}(Q_{\text{Omni}}(W_F^l), X_{l})||^2_{2}
\end{equation}
where $F_{l}(\cdot)$ represents the forward pass for a single layer $l$, $W_F^l$ represents the layer parameters and $X_{l}$ represents the layer's input. Note that the above objective is optimized independently for each of the $L$ Transformer layers.

\vspace{-2mm}
\subsection{\alg}
\label{sec:matquant}
\alg is a general purpose framework to develop a single model that can do well at any precision. It is a multi-scale training technique that works with most learning-based quantization schemes like QAT and OmniQuant discussed earlier. At its core, taking inspiration from~\citet{kusupati2022matryoshka}, \alg optimizes the quantization loss for several target bit-widths jointly. 

To have a single model for various integer precisions, we nest smaller bit-widths into large ones -- leveraging the inherent Matryoshka nature of the integer data type. So, if we want to extract a $r$-bit model from a $c$-bit model ($0<r<c$), we can just \textit{slice out} the $r$ most significant bits (MSBs) -- using a right shift, followed by a left shift of the same order. Formally, the $S(q^c, r)$ operator slices the most significant $r$ bits from a $c$-bit quantized vector $q^c$:
\begin{equation}
\label{eqn:slicing}
   S(q^c, r) = \text{clamp}\left(\left\lfloor \frac{q^c}{2^{c - r}} \right\rceil, 0, 2^r - 1  \right)  * 2^{c - r}
   \vspace{-3mm}
\end{equation}

Note that $\text{clamp}(\cdot)$ is required to curtail overflows generated by rounding. More details can be found in Appdendix~\ref{app:slicing}. Once we have this structure, we can optimize for several precisions by slicing the MSBs from the largest bit-width we are optimizing for. Let $R = \{r_1, r_2, . . ., r_K\}$ be the bit-widths we want to optimize for, $Q(\cdot,)$ represent the quantization function of the base algorithm (i.e., any learning-based quantization scheme), $\mathcal{L}(\cdot)$ represent the loss function pertaining to the base algorithm, $F(\cdot)$ represent the forward pass required to compute the loss, $\theta$ represent the set of model/auxiliary parameters we are optimizing for and let $W_F$ represent the model parameters. \alg's overall objective can be formulated as follows:
\begin{equation}
\label{eqn:matquant}
    \min_{P} \frac{1}{N} \sum_{i \in [N]} \sum_{r \in R} \lambda_{r}\cdot \mathcal{L}\left(F(S(Q(\theta, c), r), x^{\prime}_i), y^{\prime}_{i}\right)
\end{equation}
where $y^{\prime}_{i} = y_i$ for QAT and $y^{\prime}_{i} =  F_l(W^l_F, X_l^i)$ for OmniQuant, and $x^{\prime}_i = x_i$ for QAT and $x^{\prime}_i = X_l^i$ for OmniQuant. $\lambda_{r}$ is the loss reweighing factor for bit-width $r$.

In this work, we default to training \alg with three bit-widths, $R = \{8, 4, 2\}$, and subsequently perform a linear search over  $\lambda_{r}$. This process aims to optimize performance such that the model performs well across all targeted precision levels. Further, while the focus of this paper is primarily on integer data types, we discuss the possibility of extending \alg to floating-point representations in Section~\ref{sec:fp}.

A key point to note is that \alg primarily alters the quantized weight distributions across precision levels compared to the base quantization algorithm (OmniQuant or QAT). Figure~\ref{fig:teaser}c illustrates the differences in the quantized weight histograms obtained with and without \alg on Gemma-2 9B using OmniQuant. Upon close observation, we find that all the distributions of \alg are shifted to the right; that is, weights quantized with \alg tend to use more higher-valued weights. While this might not significantly impact int8 or even int4 models, int2 models benefit from utilizing more of the possible quantized weights compared to the baseline. Because int2 favors higher-valued weights, this effect propagates to higher-valued weights for int4, and then to int8. This observation highlights the potential overparameterization and freedom in the int8 data type to accommodate the more stringent needs of int2 during joint training. We further explore the effects of this phenomenon in Section~\ref{sec:spmatquant} to develop a better standalone quantization technique for a single target precision.

\begin{table*}[!ht]
\centering
\vspace{-3mm}
\caption{\alg with OmniQuant across Gemma-2 2B, 9B and Mistral 7B models. \alg performs on par with the baseline for int4 and int8 while significantly outperforming it for int2. Even the int3, int6 models obtained for free through interpolation from \alg perform comparably to the explicitly trained baselines. Task Avg. is average accuracy on the evaluation tasks ($\uparrow$) while log pplx (perplexity) is computed on C4 validation set ($\downarrow$).} 
\resizebox{1.5\columnwidth}{!}{ 
\begin{tabular}{@{}cccccccc@{}}
\toprule
Data type              & Method               & \multicolumn{2}{c}{Gemma-2 2B} & \multicolumn{2}{c}{Gemma-2 9B} & \multicolumn{2}{c}{Mistral 7B} \\\midrule
\multicolumn{1}{l}{}   & \multicolumn{1}{c}{OmniQuant} & Task Avg.       & log pplx.      & Task Avg.       & log pplx.      & Task Avg.       & log pplx.      \\\midrule
bfloat16               & \multicolumn{1}{l}{} & $68.21$       & $2.551$        & $74.38$       & $2.418$        & $73.99$       & $2.110$         \\\midrule
\multirow{2}{*}{int8} & Baseline            & $68.25$       & $2.552$        & $74.59$       & $2.418$        & $73.77$       & $2.110$         \\
                       & \alg            & $68.02$       & $2.570$         & $74.05$       & $2.438$        & $73.65$       & $2.125$        \\\midrule
\multirow{3}{*}{int4} 
                       & Sliced int8   & $62.87$       & $2.730$        & $72.26$       & $2.480$        & $38.51$       & $4.681$        \\
                       & Baseline            & $67.03$       & $2.598$        & $74.33$       & $2.451$        & $73.62$       & $2.136$        \\
                       & \alg             & $66.58$       & $2.618$        & $73.83$       & $2.491$         & $73.06$       & $2.153$        \\\midrule
\multirow{3}{*}{int2} 
                       & Sliced int8  & $39.78$       & $17.030$       & $38.11$       & $15.226$       & $37.29$       & $11.579$       \\
                       & Baseline            & $51.33$       & $3.835$        & $60.24$       & $3.292$        & $59.74$       & $3.931$        \\
                       & \alg             & $\bf52.37$         & $\bf3.800$        & $\bf63.35$       & $\bf3.187$        & $\bf62.75$       & $\bf3.153$       \\\midrule\midrule
\multirow{3}{*}{int6} 
                       & Sliced int8  & $67.72$       & $2.497$        & $74.64$       & $2.353$        & $73.00$        & $2.071$        \\
                       & Baseline            & $68.06$       & $2.554$        & $74.23$       & $2.420$         & $74.10$        & $2.112$        \\
                       & \alg             & $67.52$       & $2.574$        & $73.92$        & $2.440$        & $73.63$       & $2.127$       \\\midrule
\multirow{3}{*}{int3} 
                       & Sliced int8  & $41.35$          & $6.024$        & $54.18$       & $3.977$         & $39.21$        & $10.792$        \\
                       & Baseline            & $64.37$       & $2.727$        & $73.23$       &    $2.549$        & $71.68$       & $2.211$        \\
                       & \alg             & $64.47$       & $2.618$        & $72.87$       & $2.607$        & $71.16$       & $2.238$        \\
 \bottomrule
\end{tabular}
\label{tab:omniquant-ffn}
}
\vspace{-4mm}
\end{table*}
\subsubsection{Interpolative Behavior}
\label{sec:interpolate}

\paragraph{Slicing.}
Although we explicitly train \alg for three precisions (int8, int4, int2), we find that the resulting model, when quantized to interpolated bit-widths like int6 \& int3 by slicing (Eq.~\ref{eqn:slicing}) the int8 model, performs on par with a baseline trained explicitly for that precision. It is also significantly better than slicing an int8 quantized model. We attribute this strong interpolation in bit-width space to \alg, and present more results in Sections~\ref{sec:exp-omniquant} \& \ref{sec:exp-qat}.

\vspace{-4mm}
\paragraph{Mix'n'Match.}
\alg also enables the use of different precisions at different layers through layer-wise Mix'n'Match~\citep{devvrit2023matformer}, even though we never trained for these combinatorial possibilities. These large number of models, obtained at no cost, densely span the accuracy-vs-memory trade-off. We explore several Mix'n'Match strategies and find that having a higher precision (int8) in the middle layers and a lower precision (int2) at the start and end is the most optimal among hundreds of possible models. See Section~\ref{sec:exp-mnm} for detailed experiments.

\section{Experiments}
\label{sec:exp}
In this section, we present an empirical evaluation of \alg working with two popular learning-based quantization methods: OmniQuant (Section~\ref{sec:exp-omniquant}) and QAT (Section~\ref{sec:exp-qat}). We demonstrate \alg's efficiency on Transformer-based LLMs. Unless otherwise mentioned, our primary focus is on weight only quantization within the parameter-intensive FFN blocks of the Transformer layer.

For our experiments, we chose the default target quantization precisions to be int8, int4, and int2. Furthermore, we showcase the interpolative nature of \alg through evaluations on int6 and int3, as well as its elastic ability to densely span the accuracy-vs-cost trade-off using layer-wise Mix'n'Match (Section~\ref{sec:exp-mnm}). Finally, we ablate on improving the performance of \alg (Sections~\ref{sec:abl-weight} and \ref{sec:abl-codistill}) and extend \alg to the quantization of FFN and Attention parameters. (Section~\ref{sec:spmatquant}). 
Further training and fine-grained evaluation details are in the Appendix.

\vspace{-2mm}
\paragraph{Models and Data.} We experiment with Gemma-2~\citep{Riviere2024Gemma2I} 2B, 9B, and Mistral 7B~\citep{DBLP:mistral} models. For OmniQuant experiments, we sample 128 examples with a sequence length of 2048 from the C4 dataset~\citep{raffel2020exploring} and train using a batch size of 4. We train for a total of 10M tokens for all models except the int2 baseline, where we train the model for 20M tokens~\citep{shao2023omniquant}. For QAT experiments, we sample a fixed set of 100M tokens from the C4 dataset and train all our models using a batch size of 16 and a sequence length of 8192 for a single epoch.

\vspace{-2mm}
\paragraph{Baselines.} For OmniQuant and QAT, our primary baselines (referred to as ``Baseline'' in the tables and figures) are models trained explicitly for a given precision. When interpolating the models trained with \alg for int6 and int3, we do not perform any additional training. However, the baselines are trained explicitly for 6 and 3 bits respectively. We also compare against a sliced int8 OmniQuant/QAT baseline model to the corresponding precision (referred to as ``Sliced int8'' in the tables).


\begin{table*}[!ht]
\centering
\caption{\alg with QAT across Gemma-2 2B, 9B and Mistral 7B models. \alg performs on par with the baseline for int4 and int8 while significantly outperforming it for int2. Even the int3, int6 models obtained for free through interpolation from \alg perform comparably to the explicitly trained baselines. Task Avg. is average accuracy on the evaluation tasks ($\uparrow$) while log pplx (perplexity) is computed on C4 validation set ($\downarrow$).} 
\resizebox{1.5\columnwidth}{!}{
\begin{tabular}{@{}cccccccc@{}}
\toprule
Data type              & Method               & \multicolumn{2}{c}{Gemma-2 2B} & \multicolumn{2}{c}{Gemma-2 9B} & \multicolumn{2}{c}{Mistral 7B} \\ 
\midrule
\multicolumn{1}{l}{}   & QAT & Task Avg.      & log pplx.     & Task Avg.      & log pplx.     & Task Avg.       & log pplx.      \\
\midrule
bfloat16                & \multicolumn{1}{l}{} & $68.21$      & $2.551$       & $74.38$      & $2.418$       & $73.99$       & $2.110$         \\
\midrule
\multirow{2}{*}{\centering int8} & Baseline           & $67.82$      & $2.458$       & $74.17$      & $2.29$        & $73.48$       & $2.084$        \\
                       & \alg             & $67.44$      & $2.449$       & $74.52$      & $2.262$       & $72.58$       & $2.104$        \\ \midrule
\multirow{3}{*}{int4} 
                       & Sliced int8              & $67.13$       & $2.483$       & $73.36$      & $2.276$       & $71.76$       & $2.18$        \\ 
                       & Baseline             & $67.03$      & $2.512$       & $73.26$      & $2.324$       & $72.13$       & $2.105$        \\
                                              & \alg             & $66.59$      & $2.499$       & $73.24$      & $2.429$       & $71.99$       & $2.148$        \\\midrule
\multirow{3}{*}{int2}
                       & Sliced int8              & $39.27$      & $10.217$       & $40.40$      & $7.259$       & $37.41$        & $9.573$       \\ 
                        & Baseline             & $47.74$      & $3.433$       & $56.02$      & $2.923$       & $54.95$       & $2.699$        \\
                                              & \alg             & $\bf52.20$      & $\bf3.055$       & $\bf62.29$      & $\bf2.265$       & $\bf61.97$       & $\bf2.524$        \\\midrule\midrule
\multirow{3}{*}{int6} 
                       & Sliced int8              & $67.53$      & $2.401$       & $74.15$      & $2.232$       & $73.35$        & $2.097$ \\  
                       & Baseline             & $67.75$      & $2.460$        & $74.31$      & $2.293$       & $72.71$       & $2.077$        \\
                           & \alg             & $67.33$       & $2.453$       & $74.30$      & $2.265$       & $72.59$        & $2.106$        \\\midrule
\multirow{3}{*}{int3} 
                       & Sliced int8              & $59.56$      & $2.882$       & $68.70$      & $2.512$       & $64.33$       & $2.493$        \\
                       & Baseline             & $61.75$      & $2.678$       & $69.9$       & $2.43$        & $68.82$       & $2.197$        \\
                                              & \alg             & $60.76$      & $2.734$       & $70.41$      & $2.429$       & $67.16$       & $2.324$        \\
\bottomrule
\end{tabular}
\label{tab:qat-ffn}
}
\vspace{-5mm}
\end{table*}
\vspace{-3mm}
\paragraph{Evaluation Datasets.}
Following recent work~\citep{frantar2022gptq, ma2024affinequant}, we evaluate all the methods based on log perplexity and average zero-shot accuracy across a collection of downstream tasks. We use C4's test set to calculate perplexity, and for downstream evaluations, we test on ARC-c, ARC-e~\citep{DBLP:arc}, BoolQ~\citep{DBLP:boolqa}, HellaSwag~\citep{DBLP:hellaswag}, PIQA~\citep{DBLP:piqa}, and Winogrande~\citep{DBLP:winogrande}.

\vspace*{-2mm}
\subsection{\alg with OmniQuant}
\label{sec:exp-omniquant}
Table~\ref{tab:omniquant-ffn} shows the efficacy of \alg when used with FFN-only OmniQuant and compared to explicitly trained OmniQuant baselines for the target precisions, i.e., int8, int4, and int2, across all the models. While the average downstream accuracy of \alg for int8 and int4 quantization is within 0.5\% of the corresponding independently trained baselines, the int2 quantized models of \alg are $1.04\%$, $3.11\%$, and $3.01\%$ more accurate for Gemma-2 2B, 9B, and Mistral 7B, respectively. Similar trends and improvements follow when measuring performance through validation log perplexity. Further, the quantized int4 and int2 models \textit{sliced} from the int8 OmniQuant baseline suffer a significant drop in accuracy around int4, demonstrating that the nested structure of int8 is not well utilized.

\vspace{-2mm}
\paragraph{Sliced Interpolation.}
Beyond the target quantization granularities (int8, int4, and int2), \alg allows for bit-width interpolation to bit-widths not optimized during training. We find that the accuracy of the int6 and int3 models obtained by slicing the \alg models is comparable to their explicitly trained baselines.


\subsection{\alg with QAT}
\label{sec:exp-qat}
To further demonstrate the generality of \alg, we experiment on the same models using the popular QAT technique. Following the trend of experimental results with OmniQuant, we show in Table~\ref{tab:qat-ffn} that the models trained using \alg with QAT are comparable to the explicitly trained baselines for all the targeted bit-widths of int8 and int4. However, int2 quantized models using \alg are $4.46\%$, $6.27\%$, and $7.02\%$ more accurate for Gemma-2 2B, 9B, and Mistral 7B, respectively.

\vspace{-2mm}
\paragraph{Sliced Interpolation.}
Models trained using \alg with QAT exhibit strong interpolative performance similar to that of \alg with OmniQuant. We find that the accuracy of the int6 and int3 models obtained by \textit{slicing} the \alg models is comparable to explicitly trained baselines for both interpolated bit-widths.

While OmniQuant only trains the auxiliary parameters needed for quantization, QAT also updates the weight parameters. This potentially results in severe overfitting to the C4 subset used in the experiments. We observe this overfitting in all the experiments presented in Table~\ref{tab:qat-ffn}, where the log perplexities improve for QAT compared to OmniQuant, while the downstream accuracies suffer. This also highlights the need for high-quality data for QAT to realize its benefits; otherwise, users are better off using resource-friendly methods like OmniQuant.




\begin{figure}[!b]
  \centering
  \vspace{-5mm}
        \includegraphics[width=0.8\columnwidth]{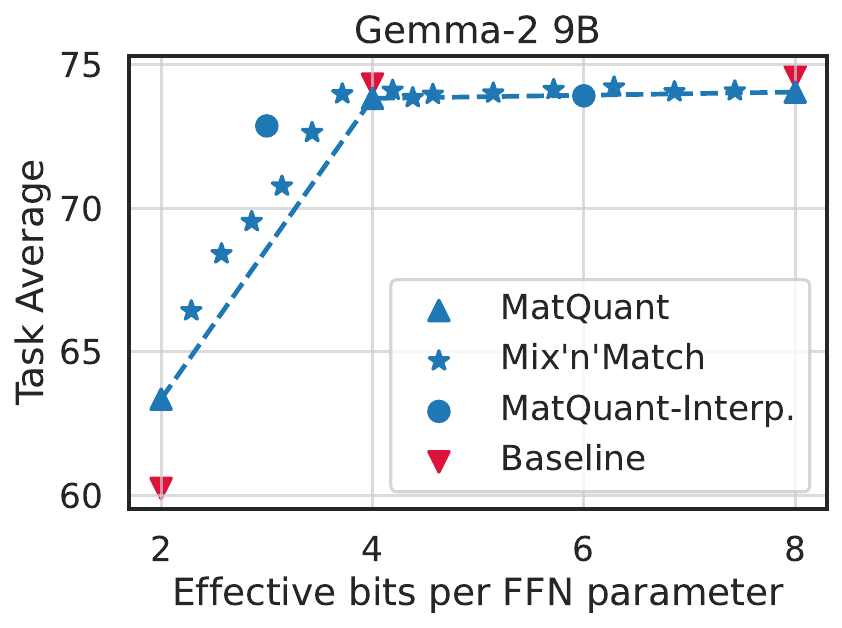}
        \vspace{-3mm}
        \caption{Mix'n'Match on Gemma-2 9B model trained using \alg with OmniQuant allows elastic accuracy-vs-cost model extraction for free during deployment.}
\label{fig:omniquant-mnm}
\end{figure}
\subsection{Layerwise Mix'n'Match}
\label{sec:exp-mnm}
Alongside the strong slicing-based interpolative properties, quantization with \alg also enables another form of elastic and interpolative behavior through Mix'n'Match. Mix'n'Match provides a mechanism to obtain a combinatorial number of strong models by using different quantization granularities, from the target bit-widths -- i.e., int8, int4, and int2 across layers. Figure~\ref{fig:omniquant-mnm} shows the ability of Mix'n'Match to densely span the accuracy-vs-bits-per-FFN-parameter (memory/cost) trade-off for the Gemma-2 9B model trained using \alg with OmniQuant. While there are many more feasible models, we only showcase the best models obtained through the strategy described in Section~\ref{sec:interpolate} and further expanded in Appendix~\ref{app:td}. Interestingly, the Mix'n'Match model, with a sub-4-bit effective width, is more accurate than the 4-bit sliced model. This opens up possibilities for effective serving depending on hardware support. Section~\ref{sec:dep} continues this discussion in greater depth.

\vspace*{-2mm}
\section{Ablations and Discussion}
\label{sec:disc}
In this section, we present design ablations to improve \alg. Section~\ref{sec:abl-weight} discusses the effect of non-uniform weighting across target precisions (int8, int4, int2), and Section~\ref{sec:abl-codistill} explores enabling co-distillation of lower precision levels (int4, int2) from the highest precision quantized model (int8). During the process of extending \alg to all Transformer parameters, not just the FFN block, we uncovered an interesting hybrid quantization algorithm (between Baseline and \alg). Section~\ref{sec:spmatquant} further details this method, called \spalg, which stabilizes the otherwise QAT baseline for all the Transformer weights. Finally, we also discuss extending \alg beyond integer data types and the considerations for effective deployment on current hardware.



\begin{table}[!t]
\centering
\caption{Design choice ablation for loss re-weighting of the 3 target bit-widths (int8, int4, int2) that \alg explicitly optimizes. Note that \alg $(0, 0, 1)$ $\equiv$ \spalg.}
\label{tab:weight}
 \resizebox{0.9\columnwidth}{!}{
\begin{tabular}{@{}ccccc@{}}
\toprule
Data type              & Weightings                                  & Gemma-2 2B & Gemma-2 9B & Mistral 7B \\\midrule
\multicolumn{1}{l}{}   & \multicolumn{1}{l}{}                                  & \multicolumn{3}{c}{Task Avg.}          \\\midrule
\multirow{4}{*}{int8} & $(0.1,0.1,1)$               & $\bf68.02$    & $\bf74.05$    & $73.27$    \\
                       & $(0.2,0.2,1)$              & $67.91$    & $73.91$    & $73.44$    \\
                       & $(0.3,0.3,1)$              & $ 68.01$    & $73.88$    & $ 73.56$    \\
                       & $(0.4,0.4,1)$              & $67.95$     & $ 73.84$    & $ \bf 73.65$    \\ \midrule
                       
\multirow{4}{*}{int4} & $(0.1,0.1,1)$               & $66.58$    & $73.83$    & $72.76$    \\
                       & $(0.2,0.2,1)$              & $ 67.47$     & $73.8$    & $73.16$    \\
                       & $(0.3,0.3,1)$                  & $66.97$    & $ 73.25$    & $ 73.47$     \\
                       & $(0.4,0.4,1)$ & $\bf 67.48$    & $\bf 74.32$    & $ \bf 73.66$    \\ \midrule
                       
\multirow{4}{*}{int2} & $(0.1,0.1,1)$               & $\bf 52.37$    & $ 63.35$    & $63.25$    \\
                       & $(0.2,0.2,1)$              & $51.88$    & $64.04$    & $ \bf 63.99$    \\
                       & $(0.3,0.3,1)$                  & $ 51.05$     & $\bf 64.1$    & $63.6$    \\
                       & $(0.4,0.4,1)$ & $51.69$    & $61.98$    & $62.75$    \\

\bottomrule
\end{tabular}
}
\end{table}
\vspace*{-3mm}
\subsection{Weightings ($\lambda_r$) for \alg}
\label{sec:abl-weight}

Depending on the constraints, we may wish to maximize the accuracy of one of the target bit-widths in \alg. Equation~\ref{eqn:matquant} provides a general formulation of \alg that supports searching over the weight $\lambda_r$ for bit-width $r$. The results in Section~\ref{sec:exp} are with the weights that have balanced performance across target precisions. Table~\ref{tab:weight} shows the weight multiplier ablation results for Gemma-2 2B, 9B, and Mistral 7B. We find that a higher relative value for $\lambda_2$ is essential in attaining good int2 performance. Increasing $\lambda_4,\lambda_8$ to improve int8 and int4 models often results in accuracy drop for the int2 models. In general, we can see that a higher relative weight for a specific precision results in increased accuracy for that bit-width. We can consider re-weighting as scaling the importance of the bits during training, and finding an optimal re-weighting recipe is an interesting research question.

\begin{table}[!t]
\centering
\caption{Design choice ablations for co-distillation within \alg. $\text{x} \rightarrow \text{y}$ represents distilling the y-bit model from the x-bit model. We note that the accuracy for int2 has significantly improved while minimally impacting the other bit-widths.}
 \resizebox{\columnwidth}{!}{
\begin{tabular}{@{}cccccc@{}}
\toprule
\multicolumn{1}{l}{}   & \multicolumn{1}{l}{Gemma-2 9B}          & \multicolumn{2}{c}{OmniQuant} & \multicolumn{2}{c}{QAT} \\ \midrule
Data type              & Config.                        & Task Avg.      & log pplx.      & Task Avg.   & log pplx.   \\ \midrule
\multirow{4}{*}{int8}  & $[8,4,2] $      & $\bf74.05$ &	$2.438$    & $74.52$ &	$2.262$ \\
&                        $[8,4,8 \rightarrow2] $      & $72.76$       & $2.473$        & $74.75$   & $2.242$      \\
                       & $[8,4,2,8 \rightarrow2] $    & $73.99$      & $\bf2.435$        & $\bf74.87$   &  $\bf2.240$   \\
                       & $[8,4,2,8 \rightarrow4;2] $ & $73.85$      & $2.437$        & $74.81$    & $2.240$     \\ \midrule
\multirow{4}{*}{int4} & 
                        $[8,4,2] $      & $\bf73.83$ &	$2.491$ & $73.24$	& $2.295$

\\ &                    $[8,4,8 \rightarrow2] $      & $72.65$      & $2.519$        & $73.76$   & $2.279$     \\
                      & $[8,4,2,8 \rightarrow2] $    & $73.63$      & $2.486$        & $73.77$   & $\bf2.276$     \\
                      & $[8,4,2,8 \rightarrow4;2] $ & $73.55$      & $\bf2.478$        & $\bf73.93$   & $2.277$      \\ \midrule
\multirow{4}{*}{int2}  & 
                         $[8,4,2] $      & $63.35$ &	$\bf3.187$ & $62.29$	& $\bf2.660$

\\ &                     $[8,4,8 \rightarrow2] $      & $62.64$       & $3.289$        & $62.31$   & $2.670$     \\
                       & $[8,4,2,8 \rightarrow2] $    & $62.91$      & $3.138$        & $\bf62.70$   &   $2.673$    \\
                       & $[8,4,2,8 \rightarrow4;2] $ & $\bf64.32$      & $3.227$        & $62.60$   & $2.670$     \\ 
\bottomrule 
\end{tabular}
}
\vspace{-6mm}
\label{tab:codistill}

\end{table}


\vspace*{-2mm}
\subsection{Co-distillation for  \alg}
\label{sec:abl-codistill}
Given the nested nature of the models trained using \alg, we explored co-distillation, where the outputs from a higher-precision model are used as the target for the lower-precision nested model, either in a standalone fashion or alongside the ground truth target (weighted equally). Table~\ref{tab:codistill} shows the effects of co-distillation applied to \alg with both OmniQuant and QAT on Gemma-2 9B. While int8 and int4 show no significant improvement, the nested int2 model benefits substantially from the int8 supervision, reaching $0.97\%$ higher accuracy than the non-co-distilled \alg with OmniQuant. Co-distillation in \alg opens up avenues for interesting design choices that can further leverage the inherent nested structure of integer data types.

\begin{table}[!b]
\centering
\caption{\spalg significantly improves upon the baseline for int2 and, at times, outperforms \alg. Crucially, int8 and int4 performances of \spalg experience a significant accuracy decrease (as shown in Tables~\ref{tab:sp_9B_omni} \& \ref{tab:sp_9B_qat}) in Appendix~\ref{app:spalg}).}
\label{tab:spmatquant_ffn}

 \resizebox{\columnwidth}{!}{
\begin{tabular}{@{}ccccccc@{}}
\toprule
int2          & \multicolumn{2}{c}{Gemma-2 2B} & \multicolumn{2}{c}{Gemma-2 9B} & \multicolumn{2}{c}{Mistral 7B} \\ \midrule
Method         & Task Avg.       & log pplx.      & Task Avg.       & log pplx.      & Task Avg.       & log pplx.      \\ \midrule
OmniQuant      & $51.33$       & $3.835$         & $60.24$       & $3.292$        & $59.74$       & $3.931$        \\

S.P. \alg  & $\bf53.42$       & $\bf3.631$        & $\bf64.02$       & $\bf3.171$        & $\bf63.58$       & $\bf2.976$        \\ 
 \alg             & $52.37$         & $3.800$        & $63.35$       & $3.187$        & $62.75$       & $3.153$        \\\midrule\midrule
QAT            & $47.74$       & $3.433$        & $56.02$       & $2.923$        & $54.95$       & $2.699$        \\

S.P. \alg      & $52.08$       & $\bf3.054$         & $\bf62.66$       & $\bf2.656$        & $61.48$       & $\bf2.509$        \\ 
\alg      & $\bf52.20$       & $3.055$        & $62.29$       & $2.660$        & $\bf61.97$       & $2.524$       \\\bottomrule
\end{tabular}
}
\end{table}
\vspace*{-2mm}
\subsection{\spalg}
\label{sec:spmatquant}
In Tables~\ref{tab:omniquant-ffn} and~\ref{tab:qat-ffn}, \alg performs on par with the explicitly trained baselines for int4, int8, and the interpolated int3 and int6 precisions. However, the int2 models show a significant accuracy improvement. To investigate this, we conducted a simple ablation in \alg by removing the loss terms for int4 and int8 (i.e., $R = \{2\}$ in Equation~\ref{eqn:matquant} or setting $\lambda_4=\lambda_8=0$) and present the results in Table~\ref{tab:spmatquant_ffn}. We call this version of \alg as \spalg. With \spalg, we observe a further boost of up to $1.05\%$, in the accuracy of int2 models at a $\sim$2\% accuracy drop in the corresponding int4 and int8 models -- int2 is still nested within int8. This improvement likely stems from the six additional bits available during \alg-style training to optimize the int2 representation. 

In the case of \spalg, gradient descent is free to tune these six additional bits to improve the overall quality of the int2 model. In \alg, since we have additional losses to preserve the performance of the int4 and int8, the int2 performance is slightly worse than  \spalg. However, since the int4 and int8 models are typically very close in accuracy to the bfloat16 model, \alg can shift some of the weights to improve the int2 model. As int4 and int8 models have substantially more quantized buckets than int2, we hypothesize that shifting some weights into adjacent buckets may not significantly affect their performance; however, it can significantly impact int2's performance. In fact, in the weight distributions presented in Fig~\ref{fig:teaser}c, we observe that \alg results in a model where larger number of weights are assigned to the higher-valued buckets. Conclusively, \alg and \spalg inherently seem to be a better way of performing low-bit quantization.

\begin{table}[!t]
\centering
\caption{Extending \alg with QAT to FFN + Attention parameters. Baseline QAT destabilizes for int2 and int3 but improves significantly through \alg \& \spalg.}
\label{tab:ffn_attn_qat}
 \resizebox{\columnwidth}{!}{
\begin{tabular}{@{}cccccc@{}}
\toprule
Data type              & Method               & \multicolumn{2}{c}{Gemma-2 9B} & \multicolumn{2}{c}{Mistral 7B} \\ \midrule
\multicolumn{1}{l}{}   & QAT & Task Avg.       & log pplx.      & Task Avg.       & log pplx.      \\\midrule
bfloat16                & \multicolumn{1}{l}{} & $74.38$       & $2.418$        & $73.99$       & $2.110$         \\ \midrule
\multirow{2}{*}{int8} & Baseline            & $74.61$       & $2.353$         & $73.73$       & $2.091$         \\
                       & \alg            & $74.85$       & $2.333$         & $73.88$       & $2.182$          \\\midrule
\multirow{3}{*}{int4} 
                       & Sliced int8  & $73.15$       & $2.362$         & $71.46$       & $2.290$         \\
                       & Baseline            & $72.98$       & $2.40$          & $71.87$       & $2.132$         \\
                       & \alg            & $74.01$       & $2.396$         & $71.44$        & $2.441$         \\ \midrule 
\multirow{4}{*}{int2} 
                       & Sliced int8  & $38.97$       & $23.467$        & $35.06$       & $10.640$        \\
                       & Baseline            & - & - & - & -             \\
& S.P. \alg                & $\bf45.69$       & $\bf3.780$          & $35.35$       & $7.761$         \\
                       & \alg            & $44.19$       & $3.826$         & $\bf38.36$       & $\bf10.971$         \\\midrule\midrule
\multirow{3}{*}{int6} 
                       & Sliced int8  & $74.49$       & $2.290$         & $73.61$       & $2.104$         \\
                       & Baseline            & $74.65$       & $2.357$         & $73.72$       & $2.093$         \\
                       & \alg            & $74.57$       & $2.340$         & $74.04$       & $2.161$         \\ \midrule
\multirow{4}{*}{int3} 
                      & Sliced int8       & $64.19$       & $2.895$         & $39.01$       & $6.018$         \\
                       & Baseline           & - & - & - & -            \\
& S.P. \alg                 & $67.68$       & $2.520$         & $67.59$       & $2.335$         \\
                       & \alg            & $63.63$       & $2.937$         & $40.55$       & $4.776$          \\\bottomrule 
\end{tabular}
}
\end{table}

\vspace{-3mm}
\paragraph{FFN + Attention Weight Quantization.}

We present results for FFN + Attention quantization for QAT in Table~\ref{tab:ffn_attn_qat}. For int8, int4 and the interpolated int6 model, \alg performs on par with the \textit{Baseline}. However, we found int2 and int3 to be very unstable while quantizing both, the FFN and the Attention parameters. Most recent works that do QAT for both the blocks~\cite{DBLP:efficientqat, DBLP:llmqat, DBLP:BitDistiller} either do some form of warm starting for the quantized parameters, or have additional distillation and auxiliary loss functions. In the naive setup of minimizing the loss with respect to the ground truth, we find QAT to be very unstable at lower precisions. On the other hand, both \alg and \spalg are very stable further highlighting the benefits brought by \alg style training. 

\vspace*{-3mm}
\subsection{Deployment Considerations}
\label{sec:dep}
\vspace*{-1mm}
Current hardware accelerators have native support for serving int8 and int4 quantized models. Additionally, custom-implemented CUDA kernels can can support various low-precision bit-widths, like int2 and int3~\citep{chee2024quip,frantar2022gptq}. \alg can generate a large number of models at inference time. Depending on the serving environment, we can choose between Mix'n'Match models and homogeneous sliced models. For example, suppose the serving environment has a memory constraint equivalent to an int3 model but lacks optimized support for int3, while supporting int2. In this case, a Mix'n'Match model with a small performance drop when compared to the sliced int3 model could be deployed. More generally, as depicted in Figure~\ref{fig:omniquant-mnm}, \alg densely spans the memory-versus-accuracy curve and can be leveraged to obtain performant model for several serving constraints. \alg can enable further research on hardware software co-design to effectively support elastic bit-widths on-the-fly during inference.

\vspace*{-2mm}
\subsection{Extension to Floating Point}
\label{sec:fp}
\vspace*{-1mm}
Extending \alg to floating-point representations, such as FP8 and FP4, presents significant challenges. Given that the exponent is encoded within the bit representation and contributes to the value as a power of 2 (i.e., effectively $\log_\text{2}$), slicing it results in buckets whose sizes increase exponentially, unlike the integer case, where bucket sizes are constant. For example, slicing the first two bits from int8 yields buckets of $0$, $64$, $128$, $192$. Here, the bucket size ($64$) is constant; however, this would not be the case when slicing two exponent bits from FP8. This is a promising avenue for future research that could further unlock the benefits of \alg, even during large-scale pretraining.



\vspace*{-2mm}
\begin{table*}[!ht]
\centering
\vspace{-3mm}
\caption{Results comparing \alg with \epalg for Gemma-2 2B, 9B, and Mistral 7B, with OmniQuant as the base algorithm. We find that for the 2-bit model, having an extra bucket significantly boosts the performance, however, this is not the case with the higher precisions.} 
\resizebox{\linewidth}{!}{ 
\begin{tabular}{@{}cccccccccc@{}}
\toprule
            Method               & \multicolumn{3}{c}{Gemma-2 2B} & \multicolumn{3}{c}{Gemma-2 9B} & \multicolumn{3}{c}{Mistral 7B} \\\midrule
\multicolumn{1}{c}{OmniQuant} & Avg. Bits & Task Avg.       & log pplx.      & Avg. Bits & Task Avg.       & log pplx.      & Avg. Bits & Task Avg.       & log pplx.      \\\midrule
bfloat16               & \multicolumn{1}{l}{} & $68.21$       & $2.551$    &    & $74.38$       & $2.418$   &     & $73.99$       & $2.110$         \\\midrule
 \alg    &  $8$      & $68.02$       & $2.570$    &  $8$     & $74.05$       & $2.438$    &  $8$    & $73.65$       & $2.125$        \\
\epalg       &  $8$      & $67.85$       & $2.580$    &  $8$     & $74.33$       & $2.446$    &  $8$    & $73.46$       & $2.132$        \\\midrule
\alg      &  $4$       & $66.58$       & $2.618$    &  $4$    & $73.83$       & $2.491$      &  $4$   & $73.06$       & $2.153$        \\
\epalg       &  $4.023$      & $66.54$       & $2.617$    &  $4.022$    & $74.26$       & $2.470$     &  $4.022$    & $73.13$       & $2.155$        \\\midrule
\alg        &  $2$     & $\bf52.37$         & $\bf3.800$   &  $2$     & $\bf63.35$       & $\bf3.187$     &  $2$   & $\bf62.75$       & $\bf3.153$       \\
\epalg       &  $2.052$      & $\bf55.70$         & $\bf3.355$   &  $2.050$     & $\bf68.25$       & $\bf2.823$      &  $2.051$  & $\bf65.99$       & $\bf2.569$       \\\midrule\midrule
\alg       &  $6$      & $67.52$       & $2.574$     &  $6$   & $73.92$        & $2.440$    &  $6$    & $73.63$       & $2.127$       \\
\epalg       &  $6.018$      & $68.01$       & $2.582$     &  $6.018$   & $74.50$        & $2.446$    &  $6.018$    & $73.59$       & $2.139$       \\\midrule
\alg         &  $3$    & $64.47$       & $2.618$   &  $3$     & $72.87$       & $2.607$   &  $3$     & $71.16$       & $2.238$    \\
\epalg      &  $3.031$       & $63.24$       & $2.757$    &  $3.029$    & $73.25$       & $2.535$   &  $3.030$     & $71.55$       & $2.228$        \\
 \bottomrule
\end{tabular}
\label{tab:old_omniquant}
}
\end{table*}

\section{Conclusions}
\label{sec:conc}
In this work, we presented \alg, a novel multi-scale training technique that leverages the nested structure of integer data types to simultaneously optimize model weight quantization across multiple precisions (int8, int4, and int2) within a single model. This general-purpose method, applicable to learning-based quantization techniques like OmniQuant and QAT, produces models with comparable accuracy to baselines for int8 and int4, while achieving significant improvements, up to $7\%$ for int2 models. \alg further enables bit-width interpolation and layer-wise mix-and-match for flexible accuracy-cost trade-offs, promising more efficient deployment of large models across various hardware settings. Finally, \alg also helped discover \spalg, which significantly improves standalone low-bit quantization. 

\begin{table}[!t]
\centering
\caption{Design choice ablations for co-distillation within \epalg. $\text{x} \rightarrow \text{y}$ represents distilling the y-bit model from the x-bit model. We note that the accuracy for $2.050$ avg. bits has significantly improved while minimally impacting the other bit-widths.}
 \resizebox{\columnwidth}{!}{
\begin{tabular}{@{}cccccc@{}}
\toprule
\multicolumn{1}{l}{}   & \multicolumn{1}{l}{Gemma-2 9B}         & \multicolumn{4}{c}{OmniQuant}  \\ \midrule
\multicolumn{1}{l}{}   & \multicolumn{1}{l}{}          & \multicolumn{2}{c}{\alg} & \multicolumn{2}{c}{E.P. \alg}  \\ \midrule
Avg. Bits              & Config.                        & Task Avg.      & log pplx. & Task Avg.      & log pplx.        \\ \midrule
\multirow{4}{*}{$\left(8, 8\right)$}  & $[8,4,2]$      & $74.05$ &	$2.438$ & $73.97$ &	$2.451$     \\
& $[8,4,8 \rightarrow2]$ & $72.76$       & $2.473$     & $73.40$       & $2.467$          \\
                       & $[8,4,2,8 \rightarrow2]$ & $73.99$      & $2.435$   & $73.46$      & $2.466$           \\
                       & $[8,4,2,8 \rightarrow4;2]$ & $73.85$      & $2.437$ & $73.32$      & $2.466$           \\ \midrule
\multirow{4}{*}{$\left(4, 4.022\right)$} & 
$[8,4,2]$  & $73.83$ &	$2.491$    & $73.88$ &	$2.481$ 

\\ & $[8,4,8 \rightarrow2]$ & $72.65$      & $2.519$     & $73.84$      & $2.488$        \\
                       & $[8,4,2,8 \rightarrow2]$ & $73.63$      & $2.486$   & $73.01$      & $2.495$            \\
                       & $[8,4,2,8 \rightarrow4;2]$ & $73.55$      & $2.478$ & $73.12$      & $2.518$            \\ \midrule
\multirow{4}{*}{$\left(2, 2.050\right)$}  & 
$[8,4,2]$  & $63.35$ &	$\bf3.187$    & $68.52$ &	$2.809$ 

\\ & $[8,4,8 \rightarrow2]$  & $62.64$       & $3.289$    & $69.2$       & $2.796$        \\
                       & $[8,4,2,8 \rightarrow2]$ & $62.91$      & $3.138$   & $\bf70.17$      & $\bf2.778$         \\
                       & $[8,4,2,8 \rightarrow4;2]$  & $\bf64.32$      & $3.227$ & $69.72$      & $2.804$            \\ 


\bottomrule 
\end{tabular}
}
\label{tab:codistill_old}

\end{table}

\section{Errata}
\label{sec:errata}
In the first draft of the paper, we had a bug and used the following equation to train and quantize our models:
\begin{equation}
\label{eqn:slicing_old}
   S(q^c, r) = \left(\left\lfloor \frac{q^c}{2^{c - r}} \right\rceil\right) * 2^{c - r}
\end{equation}
Equation~\ref{eqn:slicing_old} clearly allows an extra bucket to be included into the quantization range, i.e, a $r$-bit model would have $2^r + 1$ possible values instead of $2^r$. For example, consider slicing the first two MSBs from an unsigned int8 value, $234$. As per Equation~\ref{eqn:slicing}, $234$ first gets rounded to $4$, following which it gets clipped to $3$, and finally is scaled up to $3 * 64 = 192$ (Note that \alg int2 allows for $0$, $64$, $128$, $192$). However, since the clipping operation is missing in Equation~\ref{eqn:slicing_old}, $4$ is never clipped down to $3$, and $S(q^c, r)$ is now $4 * 64 = 256$ Thus, for certain int2 values in our final quantized model, we will have to store an extra bit. This is the case with int3, int4 and int6 as well where an extra bit is required to represent certain values. In Table~\ref{tab:old_omniquant}, we can see that the fraction of parameters that fall into this extra bucket is very small. However, for our 2-bit models, this additional bucket gives significant improvements in performance, for example, in Table~\ref{tab:old_omniquant} int2 Gemma-2 9B's average downstream accuracy goes up by $5$\% when trained with an additional bucket (referred to as \epalg in Table~\ref{tab:old_omniquant}). This number is further boosted to $6$\% with co-distillation, as evidenced by Table~\ref{tab:codistill_old}. We hypothesize that this additional bucket helps with capturing the outliers and thus leads to a significant performance boost. As highlighted by recent work~\citep{dettmers2023spqr, squeezellm}, it is crucial to store certain outliers full precision. Interestingly, we show that even a single bit is enough to capture several of these outliers, especially for low bit quantization. Finally, note that this performance boost is not very evident in higher precisions where there are enough buckets to account for the outliers.

\begin{figure}[!t]
  \centering
  \vspace{-2mm}
        \includegraphics[width=0.8\columnwidth]{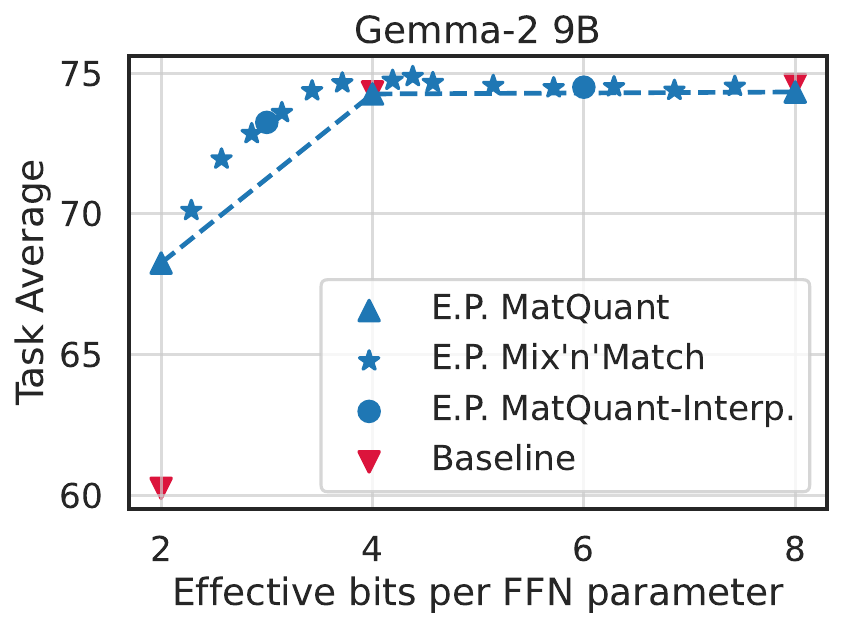}
        \vspace{-3mm}
        \caption{Mix'n'Match on Gemma-2 9B model trained using \epalg with OmniQuant as the base algorithm allows elastic pareto-optimal accuracy-vs-cost model extraction for free during deployment.}
\label{fig:omniquant-mnm-old}
\vspace*{-3mm}
\end{figure}

\paragraph{Mix'n'Match} As shown in Figure~\ref{fig:omniquant-mnm-old} with a strong int2 model (i.e., 2.050 bits on average), \epalg Mix’n’Match densely spans the Pareto-optimal accuracy-vs-bits-per-FFN-parameter (memory/cost) trade-off for Gemma-2 9B model trained using MatQuant with Omni-Quant – sometimes even improving on the bfloat16 model accuracy. Consequently, hardware supporting only int2 and int4 data types can still accommodate a model with a memory footprint similar to that of an int3 quantized model, and quality comparable or superior to int3; the additional bits required in the case of int2 can be packed into int2/int4. However, custom CUDA kernel would be required to enable sparse additions of these additional bits to the model weights.

\section*{Impact Statement}
This paper introduces a novel technique designed to advance the field of machine learning, specifically in the domain of model compression and efficient deployment for large language models. By enabling the creation of versatile, multi-scale models that can operate across various bit-widths, our work has the potential to democratize access to these powerful technologies by making them more resource-efficient and deployable on a wider range of hardware. This could lead to positive impacts such as more sustainable AI systems and greater accessibility for users with limited computational resources. While there are potential risks associated with the broad deployment of powerful AI systems, these are not unique to our work, and we believe the benefits of efficient and accessible AI through innovations like \alg have significant potential for societal good. We encourage further investigation into how novel quantization techniques can play a role in future sustainable AI development.
\section*{Acknowledgments}
We are grateful to Varun Yerram, Shreya Pathak and Devvrit for assistance in setting up inference pipelines, Shivani Agrawal. Utku Evci, Praneeth Netrapalli, Rakesh Shivanna, Tom Duerig, Abhijit Ogale, Jon Shlens, Ali Farhadi and Rahul Sukthankar
for helpful discussions, support and feedback.

\bibliography{local}

\begin{thebibliography}{40}
\providecommand{\natexlab}[1]{#1}
\providecommand{\url}[1]{\texttt{#1}}
\expandafter\ifx\csname urlstyle\endcsname\relax
  \providecommand{\doi}[1]{doi: #1}\else
  \providecommand{\doi}{doi: \begingroup \urlstyle{rm}\Url}\fi

\bibitem[Abdolrashidi et~al.(2021)Abdolrashidi, Wang, Agrawal, Malmaud, Rybakov, Leichner, and Lew]{abdolrashidi2021pareto}
A.~Abdolrashidi, L.~Wang, S.~Agrawal, J.~Malmaud, O.~Rybakov, C.~Leichner, and L.~Lew.
\newblock Pareto-optimal quantized resnet is mostly 4-bit.
\newblock In \emph{Proceedings of the IEEE/CVF Conference on Computer Vision and Pattern Recognition}, pages 3091--3099, 2021.

\bibitem[Achiam et~al.(2023)Achiam, Adler, Agarwal, Ahmad, Akkaya, Aleman, Almeida, Altenschmidt, Altman, Anadkat, et~al.]{achiam2023gpt}
J.~Achiam, S.~Adler, S.~Agarwal, L.~Ahmad, I.~Akkaya, F.~L. Aleman, D.~Almeida, J.~Altenschmidt, S.~Altman, S.~Anadkat, et~al.
\newblock Gpt-4 technical report.
\newblock \emph{arXiv preprint arXiv:2303.08774}, 2023.

\bibitem[Adelson et~al.(1984)Adelson, Anderson, Bergen, Burt, and Ogden]{adelson1984pyramid}
E.~H. Adelson, C.~H. Anderson, J.~R. Bergen, P.~J. Burt, and J.~M. Ogden.
\newblock Pyramid methods in image processing.
\newblock \emph{RCA engineer}, 29\penalty0 (6):\penalty0 33--41, 1984.

\bibitem[Adepu et~al.(2024)Adepu, Zeng, Zhang, and Singh]{adepu2024framequant}
H.~Adepu, Z.~Zeng, L.~Zhang, and V.~Singh.
\newblock Framequant: Flexible low-bit quantization for transformers.
\newblock \emph{arXiv preprint arXiv:2403.06082}, 2024.

\bibitem[Ashkboos et~al.(2024)Ashkboos, Mohtashami, Croci, Li, Jaggi, Alistarh, Hoefler, and Hensman]{quarot}
S.~Ashkboos, A.~Mohtashami, M.~L. Croci, B.~Li, M.~Jaggi, D.~Alistarh, T.~Hoefler, and J.~Hensman.
\newblock Quarot: Outlier-free 4-bit inference in rotated llms.
\newblock \emph{CoRR}, abs/2404.00456, 2024.
\newblock \doi{10.48550/ARXIV.2404.00456}.
\newblock URL \url{https://doi.org/10.48550/arXiv.2404.00456}.

\bibitem[Bengio et~al.(2013)Bengio, L{\'e}onard, and Courville]{bengio2013estimating}
Y.~Bengio, N.~L{\'e}onard, and A.~Courville.
\newblock Estimating or propagating gradients through stochastic neurons for conditional computation.
\newblock \emph{arXiv preprint arXiv:1308.3432}, 2013.

\bibitem[Bisk et~al.(2020)Bisk, Zellers, Bras, Gao, and Choi]{DBLP:piqa}
Y.~Bisk, R.~Zellers, R.~L. Bras, J.~Gao, and Y.~Choi.
\newblock {PIQA:} reasoning about physical commonsense in natural language.
\newblock In \emph{The Thirty-Fourth {AAAI} Conference on Artificial Intelligence, {AAAI} 2020, The Thirty-Second Innovative Applications of Artificial Intelligence Conference, {IAAI} 2020, The Tenth {AAAI} Symposium on Educational Advances in Artificial Intelligence, {EAAI} 2020, New York, NY, USA, February 7-12, 2020}, pages 7432--7439. {AAAI} Press, 2020.
\newblock \doi{10.1609/AAAI.V34I05.6239}.
\newblock URL \url{https://doi.org/10.1609/aaai.v34i05.6239}.

\bibitem[Chee et~al.(2024)Chee, Cai, Kuleshov, and De~Sa]{chee2024quip}
J.~Chee, Y.~Cai, V.~Kuleshov, and C.~M. De~Sa.
\newblock Quip: 2-bit quantization of large language models with guarantees.
\newblock \emph{Advances in Neural Information Processing Systems}, 36, 2024.

\bibitem[Chen et~al.(2024)Chen, Shao, Xu, Wang, Gao, Zhang, Qiao, and Luo]{DBLP:efficientqat}
M.~Chen, W.~Shao, P.~Xu, J.~Wang, P.~Gao, K.~Zhang, Y.~Qiao, and P.~Luo.
\newblock Efficientqat: Efficient quantization-aware training for large language models.
\newblock \emph{CoRR}, abs/2407.11062, 2024.
\newblock \doi{10.48550/ARXIV.2407.11062}.
\newblock URL \url{https://doi.org/10.48550/arXiv.2407.11062}.

\bibitem[Clark et~al.(2019)Clark, Lee, Chang, Kwiatkowski, Collins, and Toutanova]{DBLP:boolqa}
C.~Clark, K.~Lee, M.~Chang, T.~Kwiatkowski, M.~Collins, and K.~Toutanova.
\newblock Boolq: Exploring the surprising difficulty of natural yes/no questions.
\newblock In J.~Burstein, C.~Doran, and T.~Solorio, editors, \emph{Proceedings of the 2019 Conference of the North American Chapter of the Association for Computational Linguistics: Human Language Technologies, {NAACL-HLT} 2019, Minneapolis, MN, USA, June 2-7, 2019, Volume 1 (Long and Short Papers)}, pages 2924--2936. Association for Computational Linguistics, 2019.
\newblock \doi{10.18653/V1/N19-1300}.
\newblock URL \url{https://doi.org/10.18653/v1/n19-1300}.

\bibitem[Clark et~al.(2018)Clark, Cowhey, Etzioni, Khot, Sabharwal, Schoenick, and Tafjord]{DBLP:arc}
P.~Clark, I.~Cowhey, O.~Etzioni, T.~Khot, A.~Sabharwal, C.~Schoenick, and O.~Tafjord.
\newblock Think you have solved question answering? try arc, the {AI2} reasoning challenge.
\newblock \emph{CoRR}, abs/1803.05457, 2018.
\newblock URL \url{http://arxiv.org/abs/1803.05457}.

\bibitem[Denton et~al.(2015)Denton, Chintala, Fergus, et~al.]{denton2015deep}
E.~L. Denton, S.~Chintala, R.~Fergus, et~al.
\newblock Deep generative image models using a laplacian pyramid of adversarial networks.
\newblock \emph{Advances in neural information processing systems}, 28, 2015.

\bibitem[Dettmers et~al.(2022)Dettmers, Lewis, Belkada, and Zettlemoyer]{dettmers2022gpt3}
T.~Dettmers, M.~Lewis, Y.~Belkada, and L.~Zettlemoyer.
\newblock Gpt3. int8 (): 8-bit matrix multiplication for transformers at scale.
\newblock \emph{Advances in Neural Information Processing Systems}, 35:\penalty0 30318--30332, 2022.

\bibitem[Dettmers et~al.(2023)Dettmers, Svirschevski, Egiazarian, Kuznedelev, Frantar, Ashkboos, Borzunov, Hoefler, and Alistarh]{dettmers2023spqr}
T.~Dettmers, R.~Svirschevski, V.~Egiazarian, D.~Kuznedelev, E.~Frantar, S.~Ashkboos, A.~Borzunov, T.~Hoefler, and D.~Alistarh.
\newblock Spqr: A sparse-quantized representation for near-lossless llm weight compression.
\newblock \emph{arXiv preprint arXiv:2306.03078}, 2023.

\bibitem[Devvrit et~al.(2023)Devvrit, Kudugunta, Kusupati, Dettmers, Chen, Dhillon, Tsvetkov, Hajishirzi, Kakade, Farhadi, Jain, et~al.]{devvrit2023matformer}
F.~Devvrit, S.~Kudugunta, A.~Kusupati, T.~Dettmers, K.~Chen, I.~Dhillon, Y.~Tsvetkov, H.~Hajishirzi, S.~Kakade, A.~Farhadi, P.~Jain, et~al.
\newblock Matformer: Nested transformer for elastic inference.
\newblock \emph{arXiv preprint arXiv:2310.07707}, 2023.

\bibitem[Du et~al.(2024)Du, Zhang, Cao, Guo, Cao, Chu, and Xu]{DBLP:BitDistiller}
D.~Du, Y.~Zhang, S.~Cao, J.~Guo, T.~Cao, X.~Chu, and N.~Xu.
\newblock Bitdistiller: Unleashing the potential of sub-4-bit llms via self-distillation.
\newblock In L.~Ku, A.~Martins, and V.~Srikumar, editors, \emph{Proceedings of the 62nd Annual Meeting of the Association for Computational Linguistics (Volume 1: Long Papers), {ACL} 2024, Bangkok, Thailand, August 11-16, 2024}, pages 102--116. Association for Computational Linguistics, 2024.
\newblock \doi{10.18653/V1/2024.ACL-LONG.7}.
\newblock URL \url{https://doi.org/10.18653/v1/2024.acl-long.7}.

\bibitem[Dubey et~al.(2024)Dubey, Jauhri, Pandey, Kadian, Al-Dahle, Letman, Mathur, Schelten, Yang, Fan, et~al.]{dubey2024llama}
A.~Dubey, A.~Jauhri, A.~Pandey, A.~Kadian, A.~Al-Dahle, A.~Letman, A.~Mathur, A.~Schelten, A.~Yang, A.~Fan, et~al.
\newblock The llama 3 herd of models.
\newblock \emph{arXiv preprint arXiv:2407.21783}, 2024.

\bibitem[Frantar et~al.(2022)Frantar, Ashkboos, Hoefler, and Alistarh]{frantar2022gptq}
E.~Frantar, S.~Ashkboos, T.~Hoefler, and D.~Alistarh.
\newblock Gptq: Accurate post-training quantization for generative pre-trained transformers.
\newblock \emph{arXiv preprint arXiv:2210.17323}, 2022.

\bibitem[G~Team et~al.(2024)G~Team, Georgiev, Lei, Burnell, Bai, Gulati, Tanzer, Vincent, Pan, Wang, et~al.]{team2024gemini}
G.~G~Team, P.~Georgiev, V.~I. Lei, R.~Burnell, L.~Bai, A.~Gulati, G.~Tanzer, D.~Vincent, Z.~Pan, S.~Wang, et~al.
\newblock Gemini 1.5: Unlocking multimodal understanding across millions of tokens of context.
\newblock \emph{arXiv preprint arXiv:2403.05530}, 2024.

\bibitem[Gemma-Team(2024)]{Riviere2024Gemma2I}
Gemma-Team.
\newblock Gemma 2: Improving open language models at a practical size.
\newblock \emph{ArXiv}, abs/2408.00118, 2024.
\newblock URL \url{https://api.semanticscholar.org/CorpusID:270843326}.

\bibitem[Jacob et~al.(2018)Jacob, Kligys, Chen, Zhu, Tang, Howard, Adam, and Kalenichenko]{jacob2018quantization}
B.~Jacob, S.~Kligys, B.~Chen, M.~Zhu, M.~Tang, A.~Howard, H.~Adam, and D.~Kalenichenko.
\newblock Quantization and training of neural networks for efficient integer-arithmetic-only inference.
\newblock In \emph{Proceedings of the IEEE conference on computer vision and pattern recognition}, pages 2704--2713, 2018.

\bibitem[Jiang et~al.(2023)Jiang, Sablayrolles, Mensch, Bamford, Chaplot, de~Las~Casas, Bressand, Lengyel, Lample, Saulnier, Lavaud, Lachaux, Stock, Scao, Lavril, Wang, Lacroix, and Sayed]{DBLP:mistral}
A.~Q. Jiang, A.~Sablayrolles, A.~Mensch, C.~Bamford, D.~S. Chaplot, D.~de~Las~Casas, F.~Bressand, G.~Lengyel, G.~Lample, L.~Saulnier, L.~R. Lavaud, M.~Lachaux, P.~Stock, T.~L. Scao, T.~Lavril, T.~Wang, T.~Lacroix, and W.~E. Sayed.
\newblock Mistral 7b.
\newblock \emph{CoRR}, abs/2310.06825, 2023.
\newblock \doi{10.48550/ARXIV.2310.06825}.
\newblock URL \url{https://doi.org/10.48550/arXiv.2310.06825}.

\bibitem[Kim et~al.(2024)Kim, Hooper, Gholami, Dong, Li, Shen, Mahoney, and Keutzer]{squeezellm}
S.~Kim, C.~Hooper, A.~Gholami, Z.~Dong, X.~Li, S.~Shen, M.~W. Mahoney, and K.~Keutzer.
\newblock Squeezellm: Dense-and-sparse quantization.
\newblock In \emph{Forty-first International Conference on Machine Learning, {ICML} 2024, Vienna, Austria, July 21-27, 2024}. OpenReview.net, 2024.
\newblock URL \url{https://openreview.net/forum?id=0jpbpFia8m}.

\bibitem[Kusupati et~al.(2022)Kusupati, Bhatt, Rege, Wallingford, Sinha, Ramanujan, Howard-Snyder, Chen, Kakade, Jain, et~al.]{kusupati2022matryoshka}
A.~Kusupati, G.~Bhatt, A.~Rege, M.~Wallingford, A.~Sinha, V.~Ramanujan, W.~Howard-Snyder, K.~Chen, S.~Kakade, P.~Jain, et~al.
\newblock Matryoshka representation learning.
\newblock \emph{Advances in Neural Information Processing Systems}, 35:\penalty0 30233--30249, 2022.

\bibitem[Lin et~al.(2023)Lin, Tang, Tang, Yang, Dang, and Han]{lin2023awq}
J.~Lin, J.~Tang, H.~Tang, S.~Yang, X.~Dang, and S.~Han.
\newblock Awq: Activation-aware weight quantization for llm compression and acceleration.
\newblock \emph{arXiv preprint arXiv:2306.00978}, 2023.

\bibitem[Lin et~al.(2017)Lin, Doll{\'a}r, Girshick, He, Hariharan, and Belongie]{lin2017feature}
T.-Y. Lin, P.~Doll{\'a}r, R.~Girshick, K.~He, B.~Hariharan, and S.~Belongie.
\newblock Feature pyramid networks for object detection.
\newblock In \emph{Proceedings of the IEEE conference on computer vision and pattern recognition}, pages 2117--2125, 2017.

\bibitem[Liu et~al.(2024{\natexlab{a}})Liu, Oguz, Zhao, Chang, Stock, Mehdad, Shi, Krishnamoorthi, and Chandra]{DBLP:llmqat}
Z.~Liu, B.~Oguz, C.~Zhao, E.~Chang, P.~Stock, Y.~Mehdad, Y.~Shi, R.~Krishnamoorthi, and V.~Chandra.
\newblock {LLM-QAT:} data-free quantization aware training for large language models.
\newblock In L.~Ku, A.~Martins, and V.~Srikumar, editors, \emph{Findings of the Association for Computational Linguistics, {ACL} 2024, Bangkok, Thailand and virtual meeting, August 11-16, 2024}, pages 467--484. Association for Computational Linguistics, 2024{\natexlab{a}}.
\newblock \doi{10.18653/V1/2024.FINDINGS-ACL.26}.
\newblock URL \url{https://doi.org/10.18653/v1/2024.findings-acl.26}.

\bibitem[Liu et~al.(2024{\natexlab{b}})Liu, Zhao, Fedorov, Soran, Choudhary, Krishnamoorthi, Chandra, Tian, and Blankevoort]{spinquant}
Z.~Liu, C.~Zhao, I.~Fedorov, B.~Soran, D.~Choudhary, R.~Krishnamoorthi, V.~Chandra, Y.~Tian, and T.~Blankevoort.
\newblock Spinquant: {LLM} quantization with learned rotations.
\newblock \emph{CoRR}, abs/2405.16406, 2024{\natexlab{b}}.
\newblock \doi{10.48550/ARXIV.2405.16406}.
\newblock URL \url{https://doi.org/10.48550/arXiv.2405.16406}.

\bibitem[Ma et~al.(2024)Ma, Li, Zheng, Ling, Xiao, Wang, Wen, Chao, and Ji]{ma2024affinequant}
Y.~Ma, H.~Li, X.~Zheng, F.~Ling, X.~Xiao, R.~Wang, S.~Wen, F.~Chao, and R.~Ji.
\newblock Affinequant: Affine transformation quantization for large language models.
\newblock \emph{arXiv preprint arXiv:2403.12544}, 2024.

\bibitem[Nair and Suggala(2024)]{DBLP:cdquant}
P.~A. Nair and A.~S. Suggala.
\newblock Cdquant: Accurate post-training weight quantization of large pre-trained models using greedy coordinate descent.
\newblock \emph{CoRR}, abs/2406.17542, 2024.
\newblock \doi{10.48550/ARXIV.2406.17542}.
\newblock URL \url{https://doi.org/10.48550/arXiv.2406.17542}.

\bibitem[Raffel et~al.(2020)Raffel, Shazeer, Roberts, Lee, Narang, Matena, Zhou, Li, and Liu]{raffel2020exploring}
C.~Raffel, N.~Shazeer, A.~Roberts, K.~Lee, S.~Narang, M.~Matena, Y.~Zhou, W.~Li, and P.~J. Liu.
\newblock Exploring the limits of transfer learning with a unified text-to-text transformer.
\newblock \emph{Journal of machine learning research}, 21\penalty0 (140):\penalty0 1--67, 2020.

\bibitem[Rippel et~al.(2014)Rippel, Gelbart, and Adams]{rippel2014learning}
O.~Rippel, M.~Gelbart, and R.~Adams.
\newblock Learning ordered representations with nested dropout.
\newblock In \emph{International Conference on Machine Learning}, pages 1746--1754. PMLR, 2014.

\bibitem[Sakaguchi et~al.(2020)Sakaguchi, Bras, Bhagavatula, and Choi]{DBLP:winogrande}
K.~Sakaguchi, R.~L. Bras, C.~Bhagavatula, and Y.~Choi.
\newblock Winogrande: An adversarial winograd schema challenge at scale.
\newblock In \emph{The Thirty-Fourth {AAAI} Conference on Artificial Intelligence, {AAAI} 2020, The Thirty-Second Innovative Applications of Artificial Intelligence Conference, {IAAI} 2020, The Tenth {AAAI} Symposium on Educational Advances in Artificial Intelligence, {EAAI} 2020, New York, NY, USA, February 7-12, 2020}, pages 8732--8740. {AAAI} Press, 2020.
\newblock \doi{10.1609/AAAI.V34I05.6399}.
\newblock URL \url{https://doi.org/10.1609/aaai.v34i05.6399}.

\bibitem[Shao et~al.(2023)Shao, Chen, Zhang, Xu, Zhao, Li, Zhang, Gao, Qiao, and Luo]{shao2023omniquant}
W.~Shao, M.~Chen, Z.~Zhang, P.~Xu, L.~Zhao, Z.~Li, K.~Zhang, P.~Gao, Y.~Qiao, and P.~Luo.
\newblock Omniquant: Omnidirectionally calibrated quantization for large language models.
\newblock \emph{arXiv preprint arXiv:2308.13137}, 2023.

\bibitem[Sun et~al.(2024)Sun, Liu, Bair, and Kolter]{wanda}
M.~Sun, Z.~Liu, A.~Bair, and J.~Z. Kolter.
\newblock A simple and effective pruning approach for large language models.
\newblock In \emph{The Twelfth International Conference on Learning Representations, {ICLR} 2024, Vienna, Austria, May 7-11, 2024}. OpenReview.net, 2024.
\newblock URL \url{https://openreview.net/forum?id=PxoFut3dWW}.

\bibitem[Vaswani et~al.(2017)Vaswani, Shazeer, Parmar, Uszkoreit, Jones, Gomez, Kaiser, and Polosukhin]{vaswani2017attention}
A.~Vaswani, N.~M. Shazeer, N.~Parmar, J.~Uszkoreit, L.~Jones, A.~N. Gomez, L.~Kaiser, and I.~Polosukhin.
\newblock Attention is all you need.
\newblock In \emph{Neural Information Processing Systems}, 2017.
\newblock URL \url{https://api.semanticscholar.org/CorpusID:13756489}.

\bibitem[Xiao et~al.(2023)Xiao, Lin, Seznec, Wu, Demouth, and Han]{xiao2023smoothquant}
G.~Xiao, J.~Lin, M.~Seznec, H.~Wu, J.~Demouth, and S.~Han.
\newblock Smoothquant: Accurate and efficient post-training quantization for large language models.
\newblock In \emph{International Conference on Machine Learning}, pages 38087--38099. PMLR, 2023.

\bibitem[Yu et~al.(2019)Yu, Li, Shi, Huang, and Hua]{any_precision_dnn}
H.~Yu, H.~Li, H.~Shi, T.~S. Huang, and G.~Hua.
\newblock Any-precision deep neural networks.
\newblock \emph{ArXiv}, abs/1911.07346, 2019.
\newblock URL \url{https://api.semanticscholar.org/CorpusID:208138922}.

\bibitem[Yu et~al.(2018)Yu, Yang, Xu, Yang, and Huang]{yu2018slimmable}
J.~Yu, L.~Yang, N.~Xu, J.~Yang, and T.~Huang.
\newblock Slimmable neural networks.
\newblock \emph{arXiv preprint arXiv:1812.08928}, 2018.

\bibitem[Zellers et~al.(2019)Zellers, Holtzman, Bisk, Farhadi, and Choi]{DBLP:hellaswag}
R.~Zellers, A.~Holtzman, Y.~Bisk, A.~Farhadi, and Y.~Choi.
\newblock Hellaswag: Can a machine really finish your sentence?
\newblock In A.~Korhonen, D.~R. Traum, and L.~M{\`{a}}rquez, editors, \emph{Proceedings of the 57th Conference of the Association for Computational Linguistics, {ACL} 2019, Florence, Italy, July 28- August 2, 2019, Volume 1: Long Papers}, pages 4791--4800. Association for Computational Linguistics, 2019.
\newblock \doi{10.18653/V1/P19-1472}.
\newblock URL \url{https://doi.org/10.18653/v1/p19-1472}.

\end{thebibliography}

\newpage
\appendix
\onecolumn
\section{Particulars of the Slicing Operation.}
\label{app:slicing}
To extract a $r$-bit model from a $c$-bit model, we start by slicing out the most significant $r-1$ bits. We use $1$ for the $r^{\text{th}}$ bit if the $(r+1)^{\text{th}}$, else, we use $0$. This is captured by the round function in Equation~\ref{eqn:slicing} and is done to push values to higher buckets as we expect them to be more informative~\citep{wanda}. For example, consider the the unsigned int8 value $53$. The first two MSBs are $0$s. Naively slicing them would round down $53$ to $0$, however, we want to round it up to $1$. Since the bit corresponding to $32$ is set, i.e., the $(r+1)^{\text{th}}$ MSB, instead of rounding $53$ down to $0$, we round it up to $1$. \\
The $\text{clamp}(\cdot)$ operation is also equally important. The rounding operation in Equation~\ref{eqn:slicing} will round $240$ down to $4$, however, unsigned int2 operates with only $0, 1, 2, 3$. $\text{clamp}(\cdot)$ here would make sure that $4$ is clamped down to $3$.

\section{Addition Training Details}
\label{app:td}
We run all our experiments on TPUv5e chips. For OmniQuant experiments, we use a constant learning rate of $1e-3$ and for QAT experiments, we linearly warmup the learning rate to $1e-5$ for 150 and use a consine decay schedule thereafter. For OmniQuant experiments, we sample 128 examples with a sequence length of 2048 from the C4 dataset~\citep{raffel2020exploring} and train using a batch size of 4. We train for a total of 10M tokens for all models except the int2 baseline, where we train the model for 20M tokens~\citep{shao2023omniquant}. For Co-distillation experiments where OmniQuant is the base algorithm, we train for a total of 8.3M tokens. For QAT experiments, we sample a fixed set of 100M tokens from the C4 dataset and train all our models using a batch size of 16 and a sequence length of 8192 for a single epoch. For Attn + FFN experiments with QAT, we sample a fixed set of 300M tokens from C4 and train with a batch size of 16 for a single epoch.
We use $(\lambda_8, \lambda_4, \lambda_2) = (0.1, 0.1, 1.0)$ for all our Gemma experiments unless otherwise stated. In the case of Mistral 7B,  for OmniQuant experiments, we use $(\lambda_8, \lambda_4, \lambda_2) = (0.4, 0.4, 1.0)$, and for QAT experiments we use $(\lambda_8, \lambda_4, \lambda_2) = (0.2, 0.2, 1.0)$. For all our \epalg experiments, we use $(\lambda_8, \lambda_4, \lambda_2) = (1.0, 1.0, 1.0)$.

\paragraph{Mix'n'Match} For a fixed effective bits-per-FFN layer, where each layer was quantized to either int2, int4, or int8, we explored four different quantization strategies: Pyramid, Reverse Pyramid, Increasing, and Decreasing. In the Pyramid strategy, the initial and final layers were quantized to int2, the central layers to int8, with int4 serving as an intermediate step. The Reverse Pyramid strategy followed the opposite approach, assigning int8 to the initial and final layers, int2 to the central layers, and int4 in between. The Increasing and Decreasing strategies assigned bit precision in ascending and descending order, respectively, across the layers. Our experimental results demonstrated that, for a given effective bits per FFN layer, the Pyramid strategy consistently outperformed the others. Allocating higher precision (int8) to the middle layers helped preserve critical information, while the initial and final layers performed adequately with lower bit precision (int2 and int4), leading to a more efficient and effective quantization scheme.

\section{Detailed Downstream Evaluations for OmniQuant and QAT}
Tables~\ref{tab:down_omni_2B}, \ref{tab:down_omni_9B}, \ref{tab:down_omni_7B}, \ref{tab:down_qat_2B}, \ref{tab:down_qat_9B}, and \ref{tab:down_qat_7B} present downstream evaluation results on Gemma-2 2B, Gemma-2 9B and Mistral 7B with OmniQuant and QAT.

\section{Detailed Downstream Evaluations for \alg Re-weighting}
Tables~\ref{tab:reweighting_2B}, \ref{tab:reweighting_9B}, and \ref{tab:reweighting_7B} present downstream evaluation results for OmniQuant reweighting experiments on Gemma-2 2B, Gemma-2 9B and Mistral 7B.

\section{Detailed Downstream Evaluations for Co-Distillation}
Tables~\ref{tab:omni_codistill} and \ref{tab:qat_codistill} present the downstream evaluation and perplexity results for \alg with co-distillation on Gemma-2 9B. We present results with both, OmniQuant and QAT as the base algorithms.

\section{Detailed Evaluations for FFN + Attention Quantization}
Tables~\ref{tab:ffn_attn_9B} and \ref{tab:ffn_attn_7B} present the downstream evaluation and perplexity results for FFN + Attention quantization on Gemma-2 9B and Mistral 7B with OmniQuant and QAT.

\section{Detailed Evaluation for \spalg}
\label{app:spalg}
\begin{figure*}[htp]
\centering
\includegraphics[width=0.7\columnwidth]{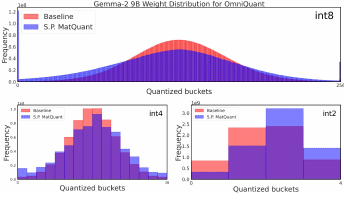}
\caption{The Figure presents the weight distribution for Gemma-2 9B when trained with \spalg for int2 quantization. The right-shifted quantized weight distribution is a consequence of \spalg's training mechanism that heavily optimizes for the first 2 MSBs of the int8 representation.}
\label{fig:spmatquant}
\end{figure*}

Tables~\ref{tab:sp_2B}, \ref{tab:sp_9B_omni}, \ref{tab:sp_9B_qat}, and \ref{tab:sp_7B} present the downstream evaluation results comparing \spalg to \alg and the \textit{Baseline} for int2 quantization of Gemma-2 2B, Gemma-2 9B and Mistral 7B with OmniQuant and QAT. Since \spalg slices 2 bits from an 8-bit model and computes loss only over the first two bits, we can evaluate the \spalg model trained for 2-bits on int4 and int8. Downstream evaluation and perplexity results for this are presented in Tables~\ref{tab:sp_9B_omni} and \ref{tab:sp_9B_qat}. We also plot the weight distribution for \spalg in Figure~\ref{fig:spmatquant}.
        



\section{Detailed Evaluation for \epalg}
Tables~\ref{tab:ep_2B_omni}, \ref{tab:ep_9B_omni}, and \ref{tab:ep_7B_omni} present downstream evaluation results for \epalg when applied to Gemma-2 2B, 9B, and Mistral 7B with OmniQuant as the base algorithm. Table~\ref{tab:ep_coditsill} presents downstream evaluation and perplexity results for our \epalg co-distillation experiments on Gemma-2 9B with OmniQuant as the base algorithm. 

\begin{table*}[h!]
\centering
\caption{Table presents the downstream evaluation results for \alg when applied to OmniQuant on Gemma-2 2B.}
\label{tab:down_omni_2B}
 \resizebox{\columnwidth}{!}{
\begin{tabular}{@{}ccccccccc@{}}
\toprule
Data type             & Method               & \multicolumn{7}{c}{Gemma-2 2B}                                           \\ \midrule
\multicolumn{1}{l}{}   & OmniQuant & ARC-c   & ARC-e   & BoolQ   & HellaSwag & PIQA    & Winogrande & Average \\\midrule
bfloat16                & \multicolumn{1}{l}{} & $50.09$ & $71.59$ & $76.45$ & $69.69$   & $78.29$ & $63.14$    & $68.21$ \\\midrule
\multirow{2}{*}{int8} & Baseline            & $50$    & $71.46$ & $76.36$ & $69.76$   & $78.24$ & $63.69$    & $68.25$ \\
                       & \alg            & $49.66$ & $71.00$ & $76.73$ & $68.85$ & $78.56$ & $63.30$ & $68.02$ \\\midrule
\multirow{3}{*}{int4} & Sliced int8  & $41.55$ & $66.12$ & $72.02$ & $62.34$ & $75.79$ & $59.43$ & $62.87$ \\
& Baseline            & $48.46$ & $70.96$ & $74.22$ & $67.66$   & $77.26$ & $63.61$    & $67.03$ \\
                       
                       & \alg            & $47.27$ & $70.79$ & $73.76$ & $66.85$ & $78.07$ & $62.75$ & $66.58$ \\
\bottomrule

\end{tabular}
}
\end{table*}
\begin{table*}[h!]
\centering
\caption{Table presents the downstream evaluation results for \alg when applied to OmniQuant on Gemma-2 9B.}
\label{tab:down_omni_9B}
 \resizebox{\columnwidth}{!}{
\begin{tabular}{@{}ccccccccc@{}}
\toprule
Data type             & Method               & \multicolumn{7}{c}{Gemma-2 9B}                                           \\ \midrule
\multicolumn{1}{l}{}   & OmniQuant & ARC-c   & ARC-e   & BoolQ   & HellaSwag & PIQA    & Winogrande & Average \\\midrule
bfloat16                & \multicolumn{1}{l}{} & $58.96$ & $77.57$ & $83.33$ & $77.31$   & $81.12$ & $67.96$    & $74.38$ \\\midrule
\multirow{2}{*}{int8} & Baseline            & $59.47$ & $77.31$ & $83.94$ & $77.35$   & $81.39$ & $68.11$    & $74.59$ \\
                       & \alg            & $57.59$ & $77.02$ & $84.01$ & $76.61$ & $81.18$ & $67.88$ & $74.05$ \\\midrule
\multirow{3}{*}{int4} 
& Sliced int8   & $55.80$ & $75.04$ & $82.32$ & $73.56$ & $80.47$ & $66.38$ & $72.26$ \\
& Baseline            & $58.79$ & $78.37$ & $83.55$ & $76.71$   & $81.45$ & $67.09$    & $74.33$ \\
                       
                       & \alg            & $58.02$ & $78.11$ & $83.24$ & $76.08$ & $80.96$ & $66.54$ & $73.83$ \\
\bottomrule

\end{tabular}
}
\end{table*}
\begin{table*}[h!]
\centering
\caption{Table presents the downstream evaluation results for \alg when applied to OmniQuant on Mistral 7B.}
\label{tab:down_omni_7B}
 \resizebox{\columnwidth}{!}{
\begin{tabular}{@{}ccccccccc@{}}
\toprule
Data type             & Method               & \multicolumn{7}{c}{Mistral 7B}                                           \\ \midrule
\multicolumn{1}{l}{}   & OmniQuant & ARC-c   & ARC-e   & BoolQ   & HellaSwag & PIQA    & Winogrande & Average \\\midrule
bfloat16                & \multicolumn{1}{l}{} & $49.57$ & $73.74$ & $84.4$  & $80.61$   & $81.18$ & $74.43$    & $73.99$ \\\midrule
\multirow{2}{*}{int8} & Baseline            & $49.23$ & $73.19$ & $83.88$ & $80.41$   & $81.39$ & $74.51$    & $73.77$ \\
                       & \alg            & $49.06$ & $72.52$ & $84.74$ & $79.21$ & $81.45$ & $74.90$ & $73.65$ \\\midrule
\multirow{3}{*}{int4} & Sliced int8  & $21.33$ & $33.67$ & $42.08$ & $28.62$ & $55.66$ & $49.72$ & $38.51$ \\
& Baseline            & $49.23$ & $73.23$ & $83.94$ & $79.9$    & $81.34$ & $74.11$    & $73.62$ \\
                       
                       & \alg            & $47.87$ & $71.55$ & $83.88$ & $78.85$ & $81.34$ & $74.90$ & $73.06$ \\
\bottomrule
\end{tabular}
}
\end{table*}
\begin{table*}[h!]
\centering
\caption{Table presents the downstream evaluation results for \alg when applied to QAT on Gemma-2 2B.}
\label{tab:down_qat_2B}
 \resizebox{\columnwidth}{!}{
\begin{tabular}{@{}ccccccccc@{}}
\toprule
Data type             & Method               & \multicolumn{7}{c}{Gemma-2 2B}                                           \\ \midrule
\multicolumn{1}{l}{}   & QAT & ARC-c   & ARC-e   & BoolQ   & HellaSwag & PIQA    & Winogrande & Average \\ \midrule
bfloat16                & \multicolumn{1}{l}{} & $50.09$ & $71.59$ & $76.45$ & $69.69$   & $78.29$ & $63.14$    & $68.21$ \\ \midrule
\multirow{2}{*}{int8} & Baseline                 & $47.78$ & $70.66$ & $75.08$ & $69.92$   & $78.35$ & $65.11$    & $67.82$ \\
                       & \alg            & $45.39$ & $71.21$ & $75.99$ & $68.74$ & $78.40$ & $64.88$ & $67.44$ \\ \midrule
\multirow{3}{*}{int4} & Sliced int8       & $46.16$ & $69.53$ & $75.35$ & $68.49$ & $78.18$ & $65.04$ & $67.13$  \\& Baseline                 & $46.16$ & $71.59$ & $73.73$ & $68.72$   & $78.62$ & $63.38$    & $67.03$ \\
                       
                       & \alg   & $44.03$ & $69.53$ & $75.84$ & $68.03$ & $77.80$ & $64.33$ & $66.59$ \\
\bottomrule
\end{tabular}
}
\end{table*}
\begin{table*}[h!]
\centering
\caption{Table presents the downstream evaluation results for \alg when applied to QAT on Gemma-2 9B.}
\label{tab:down_qat_9B}
 \resizebox{\columnwidth}{!}{
\begin{tabular}{@{}ccccccccc@{}}
\toprule
Data type             & Method               & \multicolumn{7}{c}{Gemma-2 9B}                                           \\ \midrule
\multicolumn{1}{l}{}   & QAT & ARC-c   & ARC-e   & BoolQ   & HellaSwag & PIQA    & Winogrande & Average \\ \midrule
bfloat16                & \multicolumn{1}{l}{} & $58.96$ & $77.57$ & $83.33$ & $77.31$   & $81.12$ & $67.96$    & $74.38$ \\ \midrule
\multirow{2}{*}{int8} & Baseline                  & $58.11$ & $75.38$ & $80.12$ & $78.7$    & $81.5$  & $71.19$    & $74.17$ \\
                       & \alg            & $57.68$ & $76.09$ & $82.23$ & $78.41$ & $82.26$ & $70.48$ & $74.52$ \\ \midrule
\multirow{3}{*}{int4} & Sliced int8        & $56.91$ & $75.17$ & $78.78$ & $77.02$ & $81.18$ & $71.11$ & $73.36$ \\ & Baseline                  & $56.91$ & $75.42$ & $75.38$ & $78.06$   & $81.39$ & $72.38$    & $73.26$ \\
                       
                       & \alg            & $56.66$ & $75.72$ & $77.55$ & $77.30$ & $81.23$ & $70.96$ & $73.24$ \\
\bottomrule
\end{tabular}
}
\end{table*}
\begin{table*}[h!]
\centering
\caption{Table presents the downstream evaluation results for \alg when applied to QAT on Mistral 7B.}
\label{tab:down_qat_7B}
 \resizebox{\columnwidth}{!}{
\begin{tabular}{@{}ccccccccc@{}}
\toprule
Data type             & Method               & \multicolumn{7}{c}{Mistral 7B}                                           \\ \midrule
\multicolumn{1}{l}{}   & QAT  & ARC-c   & ARC-e   & BoolQ   & HellaSwag & PIQA    & Winogrande & Average \\ \midrule
bfloat16                & \multicolumn{1}{l}{} & $49.57$ & $73.74$ & $84.4$  & $80.61$   & $81.18$ & $74.43$    & $73.99$ \\ \midrule
\multirow{2}{*}{int8} & Baseline                  & $48.89$ & $71.63$ & $82.42$ & $81.69$   & $81.18$ & $75.06$    & $73.48$ \\
                       & \alg            & $47.44$ & $71.21$ & $82.08$ & $80.31$ & $80.74$ & $73.72$ & $72.58$ \\ \midrule
\multirow{3}{*}{int4} & Sliced int8        & $47.61$ & $70.41$ & $80.21$ & $79.74$ & $79.98$ & $72.61$ & $71.76$ \\& Baseline                  & $47.27$ & $70.62$ & $81.28$ & $78.95$   & $81.12$ & $73.56$    & $72.13$ \\
                       
                       & \alg            & $45.99$ & $72.22$ & $81.90$ & $79.08$ & $80.36$ & $72.38$ & $71.99$ \\
\bottomrule
\end{tabular}
}
\end{table*}
\begin{table*}[h!]
\centering
\caption{Tables presents the downstream evaluation results on Gemma-2 2B for \alg loss reweighting when applied to OmniQuant. Weightings: $(x, y, z) \rightarrow (\lambda_8, \lambda_4, \lambda_2)$ (from Equation~\ref{eqn:matquant}). }
\label{tab:reweighting_2B}
 \resizebox{\columnwidth}{!}{
\begin{tabular}{@{}ccccccccc@{}}
\toprule
\multicolumn{1}{l}{}   & \multicolumn{1}{l}{}   & \multicolumn{7}{c}{Gemma-2 2B}                                           \\ \midrule
Data type             & Weightings   & ARC-c   & ARC-e   & BoolQ   & HellaSwag & PIQA    & Winogrande & Average \\ \midrule
\multirow{5}{*}{int8} & $(0.1,0.1,1)$ & $49.66$ & $71$    & $76.73$ & $68.85$ & $78.56$ & $63.3$  & $68.02$ \\ 
& $(0.2,0.2,1)$ & $49.4$  & $71.3$  & $76.21$ & $68.97$ & $78.29$ & $63.3$  & $67.91$ \\
& $(0.3,0.3,1)$ & $48.81$ & $71.72$ & $76.57$ & $68.95$ & $78.4$  & $63.61$ & $68.01$ \\ 
& $(0.4,0.4,1)$ & $48.72$ & $71.72$ & $76.61$ & $68.92$ & $78.73$ & $62.98$ & $67.95$ \\
& $(0.5,0.5,1)$ & $49.06$ & $71.34$ & $76.15$ & $68.86$ & $78.45$ & $62.98$ & $67.81$ \\ \midrule
\multirow{5}{*}{int4} & $(0.1,0.1,1)$ & $47.27$ & $70.79$ & $73.76$ & $66.85$ & $78.07$ & $62.75$ & $66.58$ \\
\bottomrule
\end{tabular}
}
\end{table*}

\begin{table*}[h!] 
\centering
\caption{Tables presents the downstream evaluation results on Gemma-2 9B for \alg loss reweighting when applied to OmniQuant. Weightings: $(x, y, z) \rightarrow (\lambda_8, \lambda_4, \lambda_2)$ (from Equation~\ref{eqn:matquant}). }
\label{tab:reweighting_7B}
 \resizebox{\columnwidth}{!}{
\begin{tabular}{@{}ccccccccc@{}}
\toprule
\multicolumn{1}{l}{}   & \multicolumn{1}{l}{}   & \multicolumn{7}{c}{Gemma-2 9B}                                           \\ \midrule
Data type             & Weightings   & ARC-c   & ARC-e   & BoolQ   & HellaSwag & PIQA    & Winogrande & Average \\ \midrule
\multirow{5}{*}{int8} & $(0.1,0.1,1)$ & $57.59$ & $77.02$ & $84.01$ & $76.61$ & $81.18$ & $67.88$ & $74.05$ \\
& $(0.2,0.2,1)$ & $57.76$ & $76.73$ & $83.73$ & $76.5$  & $81.34$ & $67.4$  & $73.91$ \\
& $(0.3,0.3,1)$ & $57.94$ & $76.64$ & $83.36$ & $76.56$ & $81.01$ & $67.8$  & $73.88$ \\
& $(0.4,0.4,1)$ & $58.28$ & $76.52$ & $83.15$ & $76.74$ & $80.96$ & $67.4$  & $73.84$ \\
& $(0.5,0.5,1)$ & $57.68$ & $76.68$ & $83.39$ & $76.62$ & $81.07$ & $67.09$ & $73.75$ \\ \midrule
\multirow{5}{*}{int4} & $(0.1,0.1,1)$ & $58.02$ & $78.11$ & $83.24$ & $76.08$ & $80.96$ & $66.54$ & $73.83$ \\
\bottomrule
\end{tabular}
}
\end{table*}

\begin{table*}[h!]
\centering
\caption{Tables presents the downstream evaluation results on Mistral 7B for \alg loss reweighting when applied to OmniQuant. Weightings: $(x, y, z) \rightarrow (\lambda_8, \lambda_4, \lambda_2)$ (from Equation~\ref{eqn:matquant}). }
\label{tab:reweighting_9B}
 \resizebox{\columnwidth}{!}{
\begin{tabular}{@{}ccccccccc@{}}
\toprule
\multicolumn{1}{l}{}   & \multicolumn{1}{l}{}   & \multicolumn{7}{c}{Mistral 7B}                                            \\ \midrule
Data type             & Weightings   & ARC-c   & ARC-e   & BoolQ   & HellaSwag & PIQA    & Winogrande & Average \\ \midrule
\multirow{5}{*}{int8} & $(0.1,0.1,1)$ & $49.23$ & $71.84$ & $83.94$ & $78.9$  & $81.39$ & $74.35$ & $73.27$ \\
& $(0.2,0.2,1)$ & $49.23$ & $71.97$ & $83.91$ & $79.04$ & $81.5$  & $74.98$ & $73.44$ \\
& $(0.3,0.3,1)$ & $49.32$ & $72.39$ & $84.43$ & $79.24$ & $81.23$ & $74.74$ & $73.56$ \\
& $(0.4,0.4,1)$ & $49.06$ & $72.52$ & $84.74$ & $79.21$ & $81.45$ & $74.9$  & $73.65$ \\
& $(0.5,0.5,1)$ & $49.15$ & $72.64$ & $84.65$ & $79.37$ & $81.72$ & $74.82$ & $73.72$ \\ \midrule
\multirow{5}{*}{int4} & $(0.1,0.1,1)$ & $47.61$ & $71.59$ & $83.3$  & $78.32$ & $81.61$ & $74.11$ & $72.76$ \\
\bottomrule
\end{tabular}
}
\end{table*}

\begin{table*}[t!]
\centering
\caption{Table presents the downstream evaluation and perplexity results for our \alg co-distillation experiments on Gemma-2 9B with OmniQuant.}
\label{tab:omni_codistill}
 \resizebox{\columnwidth}{!}{
\begin{tabular}{@{}cccccccccc@{}}
\toprule
\multicolumn{1}{l}{}   & OmniQuant          & \multicolumn{7}{c}{Gemma-2 9B}                                             & \multicolumn{1}{l}{} \\ \midrule
Data type             & Config.                        & ARC-c   & ARC-e   & BoolQ   & HellaSwag & PIQA    & Winogrande & Average & log pplx.            \\ \midrule
\multirow{3}{*}{int8} & $[8,4,8 \rightarrow2] $       & $57.51$ & $76.26$ & $83.30$ & $73.35$ & $80.74$ & $65.43$ & $72.76$ & $2.473$              \\
                       & $[8,4,2,8 \rightarrow2] $    & $58.19$ & $76.89$ & $83.73$ & $76.75$ & $81.39$ & $67.01$ & $73.99$ & $2.435$             \\
                       & $[8,4,2,8 \rightarrow4;2] $ & $57.68$ & $77.06$ & $83.00$ & $76.76$ & $81.45$ & $67.17$ & $73.85$ & $2.437$             \\ \midrule
\multirow{3}{*}{int4} & $[8,4,8 \rightarrow2] $       & $56.23$ & $76.47$ & $82.63$ & $73.03$ & $80.69$ & $66.85$ & $72.65$ & $2.519$             \\
                       & $[8,4,2,8 \rightarrow2] $    & $57.51$ & $76.73$ & $83.36$ & $76.23$ & $80.85$ & $67.09$ & $73.63$ & $2.486$              \\
                       & $[8,4,2,8 \rightarrow4;2] $ & $57.51$ & $76.68$ & $83.27$ & $75.85$ & $81.61$ & $66.38$ & $73.55$ & $2.478$              \\
\bottomrule
\end{tabular}
}
\end{table*}

\begin{table*}[t!]
\centering
\caption{Table presents the downstream evaluation and perplexity results for our \alg co-distillation experiments on Gemma-2 9B with QAT.}
\label{tab:qat_codistill}
 \resizebox{\columnwidth}{!}{
\begin{tabular}{@{}cccccccccc@{}}
\toprule
\multicolumn{1}{l}{}   & QAT          & \multicolumn{7}{c}{Gemma-2 9B}                                           & \multicolumn{1}{l}{} \\ \midrule
Data type             & Config.                        & ARC-c  & ARC-e  & BoolQ   & HellaSwag & PIQA    & Winogrande & Average & log pplx.            \\ \midrule
\multirow{3}{*}{int8} & $[8,4,8 \rightarrow2] $  & $57.68$ & $76.09$ & $82.60$ & $78.75$ & $82.48$ & $70.88$ & $74.75$ & $2.242$            \\
                       & $[8,4,2,8 \rightarrow2] $        & $57.76$ & $76.35$ & $81.50$ & $79.13$ & $82.43$ & $72.06$ & $74.87$ & $2.240$               \\
                       & $[8,4,2,8 \rightarrow4;2] $ & $58.19$ & $76.05$ & $81.62$ & $78.92$ & $82.21$ & $71.90$ & $74.81$ & $2.240$             \\ \midrule
\multirow{3}{*}{int4} & $[8,4,8 \rightarrow2] $     & $57.85$ & $76.81$ & $78.47$ & $77.62$ & $80.96$ & $70.88$ & $73.76$ & $2.279$                \\
                       & $[8,4,2,8 \rightarrow2] $ & $57.08$ & $75.88$ & $78.47$ & $77.65$ & $81.34$ & $72.22$ & $73.77$ & $2.276$              \\
                       & $[8,4,2,8 \rightarrow4;2] $ & $57.34$ & $75.80$ & $78.99$ & $77.67$ & $81.50$ & $72.30$ & $73.93$ & $2.277$              \\
\bottomrule
\end{tabular}
}
\end{table*}

\begin{table*}[t!]
\centering
\caption{Table presents the downstream evaluation results for \alg FFN + Attention quantization on Gemma-2 9B with QAT.}
\label{tab:ffn_attn_9B}
 \resizebox{\columnwidth}{!}{
\begin{tabular}{@{}ccccccccc@{}}
\toprule
Data type             & Method               & \multicolumn{7}{c}{Gemma-2 9B}                                           \\ \midrule
\multicolumn{1}{l}{}   & \multicolumn{1}{l}{} & ARC-c   & ARC-e   & BoolQ   & HellaSwag & PIQA    & Winogrande & Average \\\midrule
bfloat16                & \multicolumn{1}{l}{} & $58.96$ & $77.57$ & $83.33$ & $77.31$   & $81.12$ & $67.96$    & $74.38$ \\\midrule
\multirow{2}{*}{int8} & Baseline            & $58.62$ & $77.02$ & $83.43$ & $79.01$   & $81.34$ & $68.27$    & $74.61$ \\
                       & \alg            & $59.47$ & $77.99$ & $84.13$ & $77.85$ & $81.23$ & $68.43$ & $74.85$ \\\midrule
\multirow{3}{*}{int4} & Sliced int8  & $57.42$ & $76.01$ & $80.86$ & $76.34$ & $80.03$ & $68.27$ & $73.15$ \\& Baseline            & $56.06$ & $74.96$ & $79.27$ & $77.83$   & $80.25$ & $69.53$    & $72.98$ \\
                       
                       & \alg            & $58.79$ & $75.80$ & $84.89$ & $76.26$ & $81.23$ & $67.09$ & $74.01$ \\
\bottomrule 
\end{tabular}
}
\end{table*}

\begin{table*}[t!]
\centering
\caption{Table presents the downstream evaluation results for \alg FFN + Attention quantization on Mistral 7B with QAT.}
\label{tab:ffn_attn_7B}
 \resizebox{\columnwidth}{!}{
\begin{tabular}{@{}ccccccccc@{}}
\toprule
Data type             & Method               & \multicolumn{7}{c}{Mistral 7B}                                 \\ \midrule
\multicolumn{1}{l}{}   & \multicolumn{1}{l}{} & ARC-c   & ARC-e   & BoolQ   & HellaSwag & PIQA    & Winogrande & Average \\ \midrule
bfloat16                & \multicolumn{1}{l}{} & $49.57$ & $73.74$ & $84.4$  & $80.61$   & $81.18$ & $74.43$    & $73.99$ \\ \midrule
\multirow{2}{*}{int8} & Baseline            & $49.23$ & $72.9$  & $83.49$ & $80.26$   & $81.28$ & $75.22$    & $73.73$ \\
                       & \alg            & $50.09$ & $73.44$ & $83.73$ & $80.73$ & $81.39$ & $73.88$ & $73.88$ \\ \midrule
\multirow{3}{*}{int4} & Sliced int8  & $45.99$ & $71.55$ & $81.19$ & $76.90$ & $80.58$ & $72.53$ & $71.46$ \\& Baseline            & $48.04$ & $71.72$ & $78.87$ & $78.93$   & $80.36$ & $73.32$    & $71.87$ \\
                       
                       & \alg            & $46.59$ & $70.29$ & $81.65$ & $77.34$ & $80.25$ & $72.53$ & $71.44$  \\
\bottomrule
\end{tabular}
}
\end{table*}
\begin{table*}[t!]
\centering  
\caption{Table presents downstream evaluation and perplexity results for \spalg, comparing it with \alg and the  \textit{Baseline} for int2 quatization of Gemma-2 2B with OmniQuant and QAT.}
\label{tab:sp_2B}
 \resizebox{\columnwidth}{!}{
\begin{tabular}{@{}cccccccccc@{}}\toprule
\multicolumn{1}{l}{}       & int2      & \multicolumn{7}{c}{Gemma2-2B}                                            \\ \midrule
\multicolumn{1}{l}{}       & Method     & ARC-c   & ARC-e   & BoolQ   & HellaSwag & PIQA    & Winogrande & Task Avg. & $\log$ pplx. \\\midrule
\multirow{3}{*}{OmniQuant} & S.P. \alg  & $29.78$ & $57.70$	& $63.39$ &	$44.32$	& $68.66$ &	$56.67$	& $53.42$	&	$3.631$ \\
                           & Baseline   & $31.31$ & $53.58$ & $62.2$  & $40.78$   & $66.05$ & $54.06$    & $51.33$ &  $3.835$\\
                           & \alg  & $29.95$ &	$54.21$	& $64.40$ &	$44.37$ &	$66.81$ &	$54.46$ &	$52.37$	& $3.800$ \\ \midrule
\multirow{3}{*}{QAT}       & S.P. \alg  & $28.07$ & $52.36$ & $62.87$ & $46.80$ & $68.88$ & $53.51$ & $52.08$ & $3.054$ \\
                           & Baseline   & $24.66$ & $43.22$ & $62.17$ & $38.39$   & $64.42$ & $53.59$    & $47.74$ & $3.433$ \\
                           & \alg  & $28.33$ & $51.85$ & $63.64$ & $46.94$ & $68.28$ & $54.14$ & $52.20$ & $3.055$ \\\bottomrule
\end{tabular}
}
\end{table*}

\begin{table*}[t!]
\centering
\caption{Table presents downstream evaluation and perplexity results for \spalg, comparing it with \alg and the  \textit{Baseline} for int2, int4, int8 quatization of Gemma-2 9B with Baseline. Note that the model was trained with \spalg for int2; the int4 and int8 model were sliced post training.}
\label{tab:sp_9B_omni}
 \resizebox{\columnwidth}{!}{
\begin{tabular}{@{}cccccccccc@{}}\toprule
\multicolumn{1}{l}{}  & \multicolumn{1}{l}{} &
\multicolumn{8}{c}{Gemma-2 9B}                                                       \\ \midrule
Data type             & Method               & ARC-c   & ARC-e   & BoolQ   & HellaSwag & PIQA    & Winogrande & Average & $\log$ pplx. \\ \midrule
\multirow{3}{*}{int8} & S.P. \alg      & $57.94$ &	$76.64$ & $82.66$ &	$76.98$ & $81.01$ &	$67.56$ & $73.80$	& $2.372$   \\
                      & Baseline            & $59.47$ & $77.31$ & $83.94$ & $77.35$   & $81.39$ & $68.11$    & $74.59$ & $2.418$   \\
                      & \alg             & $57.59$ & $77.02$ & $84.01$ & $76.61$ & $81.18$ & $67.88$ & $74.05$ & $2.438$   \\ \midrule
\multirow{3}{*}{int4} & S.P. \alg          & $57.17$	& $76.39$	& $81.47$	& $75.81$ &	$80.85$ &	$66.38$ & 	$73.01$ &	$2.420$   \\
                      & Baseline            & $58.79$ & $78.37$ & $83.55$ & $76.71$   & $81.45$ & $67.09$    & $74.33$ & $2.451$   \\
                      & \alg              & $58.02$ & $78.11$ & $83.24$ & $76.08$ & $80.96$ & $66.54$ & $73.83$ & $2.491$   \\ \midrule
\multirow{3}{*}{int2} & S.P. \alg           & $40.44$ & $66.75$ & $77.92$ & $60.42$ & $72.52$ & $66.06$ & $64.02$ & $3.171$   \\
                      & Baseline            & $39.16$ & $63.43$ & $72.11$ & $52.24$   & $72.63$ & $61.88$    & $60.24$ & $3.292$   \\
                      & \alg              & $40.78$ & $67.85$ & $73.64$ & $60.56$ & $72.09$ & $65.19$ & $63.35$ & $3.187$
                      \\ \bottomrule
\end{tabular}
}
\end{table*}
\begin{table*}[t!]
\centering
\caption{Table presents downstream evaluation and perplexity results for \spalg, comparing it with \alg and the  \textit{Baseline} for int2, int4, int8 quatization of Gemma-2 9B with Baseline. Note that the model was trained with \spalg for int2; the int4 and int8 model were sliced post training.}
\label{tab:sp_9B_qat}
 \resizebox{\columnwidth}{!}{
\begin{tabular}{cccccccccc}\toprule
\multicolumn{1}{l}{}  & \multicolumn{1}{l}{} & \multicolumn{8}{c}{Gemma-2 9B}                                                       \\ \midrule
Data type             & Method               & ARC-c   & ARC-e   & BoolQ   & HellaSwag & PIQA    & Winogrande & Average & log pplx. \\ \midrule
\multirow{3}{*}{int8} & S.P. \alg          & $55.89$ & $75.84$ & $79.57$ & $75.47$ & $81.07$ & $68.43$ & $72.71$ & $2.363$   \\
                      & Baseline                  & $58.11$ & $75.38$ & $80.12$ & $78.7$ & $81.5$ & $71.19$ & $74.17$ & $2.29$    \\
                      & MatQuant             & $57.68$ & $76.09$ & $82.23$ & $78.41$ & $82.26$ & $70.48$ & $74.52$ & $2.262$   \\ \midrule
\multirow{3}{*}{int4} & S.P. \alg          & $54.95$ & $75.59$ & $75.05$ & $74.60$ & $80.79$ & $69.06$ & $71.67$ & $2.394$   \\
                      & Baseline                  & $56.91$ & $75.42$ & $75.38$ & $78.06$ & $81.39$ & $72.38$ & $73.26$ & $2.324$   \\
                      & MatQuant             & $56.66$ & $75.72$ & $77.55$ & $77.30$ & $81.23$ & $70.96$ & $73.24$ & $2.295$   \\ \midrule
\multirow{3}{*}{int2} & S.P. \alg          & $40.53$ & $67.38$ & $66.91$ & $63.62$ & $75.63$ & $61.88$ & $62.66$ & $2.656$   \\
                      & Baseline                  & $33.45$ & $55.43$ & $62.26$ & $54.8$    & $70.51$ & $59.67$    & $56.02$ & $2.923$   \\
                      & MatQuant             & $41.21$ & $66.84$ & $65.41$ & $63.61$ & $75.41$ & $61.25$ & $62.29$ & $2.660$  \\\bottomrule 
\end{tabular}
}
\end{table*}

\begin{table*}[t!]
\centering 
\caption{Table presents downstream evaluation and perplexity results for \spalg, comparing it with \alg, and the \textit{Baseline} for int2 quatization of Mistral 7B. Results are presented for both, OmniQuant and QAT as the base algorithms.}
\label{tab:sp_7B}
 \resizebox{\columnwidth}{!}{
\begin{tabular}{@{}cccccccccc@{}} \midrule
\multicolumn{1}{l}{}       & int2      & \multicolumn{7}{c}{Mistral 7B}                                           \\ \midrule
\multicolumn{1}{l}{}       & Method     & ARC-c   & ARC-e   & BoolQ   & HellaSwag & PIQA    & Winogrande & Task Avg. & $\log$ pplx. \\ \midrule
\multirow{3}{*}{OmniQuant} & S.P. \alg  & $37.63$ &	$64.14$	& $72.45$ &	$67.47$ &	$74.81$ &	$64.96$ &	$63.58$ &	$2.976$ \\
                           & Baseline   & $36.69$ & $61.36$ & $70.06$ & $57.47$   & $70.67$ & $62.19$    & $59.74$ & $3.931$\\
                           & \alg   & $37.88$ &	$62.58$ & 	$73.15$	& $65.89$	& $73.88$	& $63.14$ &	$62.75$	& $3.153$ \\ \midrule
\multirow{3}{*}{QAT}       & S.P. \alg  & $35.24$ & $57.15$ & $69.88$ & $66.02$ & $75.41$ & $65.19$ & $61.48$ & $2.509$\\
                           & Baseline   & $29.78$ & $48.23$ & $64.5$  & $55.11$   & $70.84$ & $61.25$    & $54.95$ & $2.694$\\
                           & \alg   & $35.58$ & $56.36$ & $72.66$ & $66.68$ & $74.32$ & $66.22$ & $61.97$ & $2.524$ \\ \midrule
\end{tabular}
}
\end{table*}

\begin{table*}[h!]
\centering
\caption{Table presents the downstream evaluation results for \epalg when applied to OmniQuant on Gemma-2 2B.}
\label{tab:ep_2B_omni}
 \resizebox{\columnwidth}{!}{
\begin{tabular}{@{}ccccccccc@{}}
\toprule
Avg. Bits             & Method               & \multicolumn{7}{c}{Gemma-2 2B}                                           \\ \midrule
\multicolumn{1}{l}{}   & OmniQuant & ARC-c   & ARC-e   & BoolQ   & HellaSwag & PIQA    & Winogrande & Average \\\midrule
bfloat16                & \multicolumn{1}{l}{} & $50.09$ & $71.59$ & $76.45$ & $69.69$   & $78.29$ & $63.14$    & $68.21$ \\\midrule
\multirow{2}{*}{} 
                       8 & \alg            & $49.66$ & $71.00$ & $76.73$ & $68.85$ & $78.56$ & $63.30$ & $68.02$ \\
                       8 & \epalg &  $48.04$ & $71.8$  & $75.78$ & $67.64$   & $78.07$ & $63.22$    & $67.42$ \\ \midrule
\multirow{3}{*}{} 
                       
                       4 & \alg            & $47.27$ & $70.79$ & $73.76$ & $66.85$ & $78.07$ & $62.75$ & $66.58$ \\ 
                       4.023 & \epalg &  $45.65$ & $70.29$ & $74.8$  & $66.07$   & $77.58$ & $62.27$    & $66.11$ \\ \midrule
\multirow{3}{*}{} 
                       
                       2 & \alg            & $29.95$ & $54.21$ & $64.40$ & $44.37$ & $66.81$ & $54.46$ & $52.37$ \\
                       2.052 & \epalg &  $34.39$ & $59.64$ & $62.69$ & $52.11$   & $69.86$ & $55.56$    & $55.71$ \\\midrule \midrule
\multirow{3}{*}{}  
                       
                       6 & \alg            & $48.89$ & $70.50$ & $75.69$ & $68.89$ & $78.40$ & $62.75$ & $67.52$ \\
                       6.018 & \epalg &  $47.1$  & $71.46$ & $76.02$ & $67.47$   & $77.91$ & $63.61$    & $67.26$ \\ \midrule
\multirow{3}{*}{} 
                       
                       3 & \alg            & $44.03$ & $67.09$ & $74.25$ & $62.78$ & $77.26$ & $61.40$ & $64.47$ \\
                       3.031 & \epalg &  $44.45$ & $68.56$ & $69.11$ & $62.28$   & $75.95$ & $62.59$    & $63.82$ \\ \bottomrule

\end{tabular}
}
\end{table*}

\begin{table*}[h!]
\centering
\caption{Table presents the downstream evaluation results for \epalg when applied to OmniQuant on Gemma-2 9B.}
\label{tab:ep_9B_omni}
 \resizebox{\columnwidth}{!}{
\begin{tabular}{@{}ccccccccc@{}}
\toprule
Avg. Bits             & Method               & \multicolumn{7}{c}{Gemma-2 9B}                                           \\ \midrule
\multicolumn{1}{l}{}   & OmniQuant & ARC-c   & ARC-e   & BoolQ   & HellaSwag & PIQA    & Winogrande & Average \\\midrule
bfloat16                & \multicolumn{1}{l}{} & $58.96$ & $77.57$ & $83.33$ & $77.31$   & $81.12$ & $67.96$    & $74.38$ \\\midrule
\multirow{2}{*}{} 
                       8 & \alg            & $57.59$ & $77.02$ & $84.01$ & $76.61$ & $81.18$ & $67.88$ & $74.05$ \\
                       8 & \epalg & $58.11$ & $78.03$ & $83.27$ & $76.17$   & $81.18$ & $67.09$    & $73.97$ \\\midrule
\multirow{3}{*}{}

                       4 & \alg            & $58.02$ & $78.11$ & $83.24$ & $76.08$ & $80.96$ & $66.54$ & $73.83$ \\ 
                       4.022 & \epalg & $57.25$ & $77.36$ & $84.86$ & $75.52$   & $81.5$  & $66.77$    & $73.88$ \\ \midrule
\multirow{3}{*}{}

                       2 & \alg            & $40.78$ & $67.85$ & $73.64$ & $60.56$ & $72.09$ & $65.19$ & $63.35$ \\
                       2.050 & \epalg & $48.72$ & $72.18$ & $79.2$  & $68.11$   & $76.17$ & $66.77$    & $68.52$ \\ \midrule \midrule
 \multirow{3}{*}{} 
                       
                       6 & \alg            & $57.25$ & $76.94$ & $84.04$ & $76.63$ & $81.34$ & $67.32$ & $73.92$ \\
                       6.018 & \epalg & $58.87$ & $78.03$ & $83.61$ & $76.18$   & $81.45$ & $67.09$    & $74.21$ \\ \midrule
\multirow{3}{*}{} 
                       
                       3 & \alg            & $55.80$ & $76.89$ & $81.99$ & $74.27$ & $80.14$ & $68.11$ & $72.87$ \\
                       3.029 & \epalg & $55.46$ & $76.14$ & $84.04$ & $74.49$   & $80.14$ & $67.32$    & $72.93$ \\ \bottomrule

\end{tabular}
}
\end{table*}

\begin{table*}[h!]
\centering
\caption{Table presents the downstream evaluation results for \epalg when applied to OmniQuant on Mistral 7B.}
\label{tab:ep_7B_omni}
 \resizebox{\columnwidth}{!}{
\begin{tabular}{@{}ccccccccc@{}}
\toprule
Data type             & Method               & \multicolumn{7}{c}{Mistral 7B}                                           \\ \midrule
\multicolumn{1}{l}{}   & OmniQuant & ARC-c   & ARC-e   & BoolQ   & HellaSwag & PIQA    & Winogrande & Average \\\midrule
bfloat16                & \multicolumn{1}{l}{} & $49.57$ & $73.74$ & $84.4$  & $80.61$   & $81.18$ & $74.43$    & $73.99$ \\\midrule
\multirow{2}{*}{} 
                       8 & \alg            & $49.06$ & $72.52$ & $84.74$ & $79.21$ & $81.45$ & $74.90$ & $73.65$ \\
                       8 & \epalg & $48.04$ & $73.44$ & $84.13$ & $79.37$   & $81.12$ & $74.66$    & $73.46$ \\ \midrule
\multirow{3}{*}{} 
                       
                       4 & \alg            & $47.87$ & $71.55$ & $83.88$ & $78.85$ & $81.34$ & $74.90$ & $73.06$ \\
                       4.022 & \epalg & $48.21$ & $72.69$ & $83.49$ & $78.82$   & $81.12$ & $74.43$    & $73.13$ \\ \midrule
\multirow{3}{*}{}

                       2 & \alg            & $37.88$ & $62.58$ & $73.15$ & $65.89$ & $73.88$ & $63.14$ & $62.75$ \\
                       2.051 & \epalg & $41.38$ & $67.42$ & $71.62$ & $71.98$   & $77.86$ & $65.67$    & $65.99$ \\ \midrule \midrule
\multirow{3}{*}{} 
                       
                       6 & \alg            & $49.40$ & $72.47$ & $84.68$ & $79.52$ & $81.34$ & $74.35$ & $73.63$ \\ 
                       6.018 & \epalg & $48.46$ & $72.98$ & $84.07$ & $79.64$   & $81.18$ & $75.22$    & $73.59$ \\
\midrule
\multirow{3}{*}{} 
                       
                       3 & \alg            & $47.35$ & $71.00$ & $80.00$ & $76.96$ & $80.30$ & $71.35$ & $71.16$ \\ 
                       3.030 & \epalg & $45.65$ & $71.21$ & $80.43$ & $78.31$   & $81.07$ & $72.61$    & $71.55$ \\ \bottomrule
\end{tabular}
}
\end{table*}


\begin{table*}[t!]
\centering
\caption{Table presents the downstream evaluation and perplexity results for our \epalg co-distillation experiments on Gemma-2 9B with OmniQuant.}
\label{tab:ep_coditsill}
 \resizebox{\columnwidth}{!}{
\begin{tabular}{@{}cccccccccc@{}}
\toprule
\multicolumn{1}{l}{}   & OmniQuant          & \multicolumn{7}{c}{Gemma-2 9B}                                             & \multicolumn{1}{l}{} \\ \midrule
Avg. Bits             & Config.                        & ARC-c   & ARC-e   & BoolQ   & HellaSwag & PIQA    & Winogrande & Average & log pplx.            \\ \midrule
\multirow{3}{*}{$8$} & $[8,4,8 \rightarrow2]$       & $57.59$ & $77.27$ & $81.83$ & $75.48$   & $81.01$ & $67.25$    & $73.4$  & $2.467$              \\
                       & $[8,4,2,8 \rightarrow2]$    & $57.17$ & $77.36$ & $82.2$  & $75.82$   & $80.96$ & $67.25$    & $73.46$ & $2.466$              \\
                       & $[8,4,2,8 \rightarrow4;2]$ & $56.4$  & $77.82$ & $82.32$ & $75.02$   & $80.63$ & $67.72$    & $73.32$ & $2.466$              \\ \midrule
\multirow{3}{*}{$4.022$} & $[8,4,8 \rightarrow2]$       & $57.68$ & $78.45$ & $82.97$ & $75.5$    & $80.85$ & $67.56$    & $73.84$ & $2.488$              \\
                       & $[8,4,2,8 \rightarrow2]$    & $57.51$ & $77.61$ & $80.46$ & $74.74$   & $81.12$ & $66.61$    & $73.01$ & $2.495$              \\
                       & $[8,4,2,8 \rightarrow4;2]$ & $56.57$ & $77.99$ & $82.54$ & $74.77$   & $80.58$ & $66.3$     & $73.12$ & $2.518$              \\
\bottomrule
\end{tabular}
}
\end{table*}

\begin{table}[!b]
\centering
\caption{Table presents downstream task average and log pplx (perplexity) when applied to OmniQuant and QAT on Gemma-2 2B, 9B and Mistral 7B models.}

 \resizebox{\columnwidth}{!}{
\begin{tabular}{@{}ccccccc@{}}
\toprule
int2          & \multicolumn{2}{c}{Gemma-2 2B} & \multicolumn{2}{c}{Gemma-2 9B} & \multicolumn{2}{c}{Mistral 7B} \\ \midrule
Method         & Task Avg.       & log pplx.      & Task Avg.       & log pplx.      & Task Avg.       & log pplx.      \\ \midrule
OmniQuant      & $51.33$       & $3.835$         & $60.24$       & $3.292$        & $59.74$       & $3.931$        \\
S.P. \alg  & $53.42$       & $3.631$        & $64.02$       & $3.171$        & $63.58$       & $2.976$        \\ 
 \alg             & $52.37$         & $3.800$        & $63.35$       & $3.187$        & $62.75$       & $3.153$        \\
 S.P. E.P. \alg & $57.38$       & $3.185$        & $68.58$       & $2.857$        & $67.36$       & $2.464$        \\ 
 E.P. \alg      & $55.71$       & $3.292$        & $68.52$       & $2.809$        & $65.99$       & $2.569$        \\\midrule\midrule
QAT            & $47.74$       & $3.433$        & $56.02$       & $2.923$        & $54.95$       & $2.699$        \\
S.P. \alg      & $52.08$       & $3.054$         & $62.66$       & $2.656$        & $61.48$       & $2.509$        \\ 
\alg      & $52.20$       & $3.055$        & $62.29$       & $2.660$        & $61.97$       & $2.524$       \\
S.P. E.P. \alg      & $53.18$       & $3.090$         & $62.53$       & $2.706$        & $61.55$       & $2.435$        \\ 
E.P. \alg      & $52.43$       & $3.153$        & $62.32$       & $2.756$        & $61.29$       & $2.474$       \\\bottomrule
\end{tabular}
}
\end{table}

\end{document}